\documentclass[sigconf]{acmart} % authordraft
\AtBeginDocument{%
  }

\setcopyright{acmlicensed}
\copyrightyear{2018}
\acmYear{2018}
\acmDOI{XXXXXXX.XXXXXXX}
\acmConference[Conference acronym 'XX]{Make sure to enter the correct
  conference title from your rights confirmation email}{June 03--05,
  2018}{Woodstock, NY}
\acmISBN{978-1-4503-XXXX-X/2018/06}

\usepackage{float}
\usepackage{multirow} 
\usepackage{graphicx}
\usepackage{tcolorbox}
\tcbuselibrary{breakable}
\newcommand{\methodname}{Pano-R1}
\begin{document}

\title{Omnidirectional Spatial Modeling from Correlated Panoramas}

\author{Xinshen Zhang}
\affiliation{%
  \institution{The Hong Kong Polytechnic University}
  \country{Hong Kong, China}
}
% \email{22099321d@gmail.com}

\author{Tongxi Fu}
\affiliation{%
  \institution{Zhejiang Sci-Tech University}
  \country{Hangzhou, China}
}
% \email{22099321d@gmail.com}

\author{Xu Zheng}
\authornote{Corresponding author}
\affiliation{%
  \institution{HKUST(GZ)}
  \country{Guangzhou, China}
}
% \email{zhengxu128@gmail.com}

% \affiliation{%

% \affiliation{%

% \affiliation{%


\renewcommand{\shortauthors}{Zhang et al.}

\begin{abstract}
Omnidirectional scene understanding is vital for various downstream applications, such as embodied AI, autonomous driving, and immersive environments, yet remains challenging due to geometric distortion and complex spatial relations in 360° imagery. Existing omnidirectional methods achieve scene understanding within a single frame while neglecting cross-frame correlated panoramas.
% benchmarks seldom address cross-frame reasoning in panoramic settings. 
To bridge this gap, we introduce \textbf{CFpano}, the \textbf{first} benchmark dataset dedicated to cross-frame correlated panoramas visual question answering in the holistic 360° scenes. CFpano consists of over 2700 images together with over 8000 question-answer pairs, and the question types include both multiple choice and open-ended VQA.
% comprising over 8,000 questions.
Building upon our CFpano, we further present \methodname, a multi-modal large language model (MLLM) fine-tuned with Group Relative Policy Optimization (GRPO) and a set of tailored reward functions for robust and consistent reasoning with cross-frame correlated panoramas. Benchmark experiments with existing MLLMs are conducted with our CFpano.
The experimental results demonstrate that \methodname achieves state-of-the-art performance across both multiple-choice and open-ended VQA tasks, outperforming strong baselines on all major reasoning categories (\textbf{+5.37\%} in overall performance). Our analyses validate the effectiveness of GRPO and establish a new benchmark for panoramic scene understanding.
\end{abstract}   




\received{20 February 2007}
\received[revised]{12 March 2009}
\received[accepted]{5 June 2009}

\maketitle

\section{Introduction}

% \begin{figure}[ht]
%   \flushright
%   \includegraphics[width=0.9\columnwidth]{Figures/model_performance_radar_combined.pdf}
%   \caption{Performance comparison of models on our benchmark.}
%   \label{fig:radar}
% \end{figure}

\begin{figure}[ht]  % 或 [H] 如果需要固定位置
\flushright
\includegraphics[width=0.98\columnwidth]{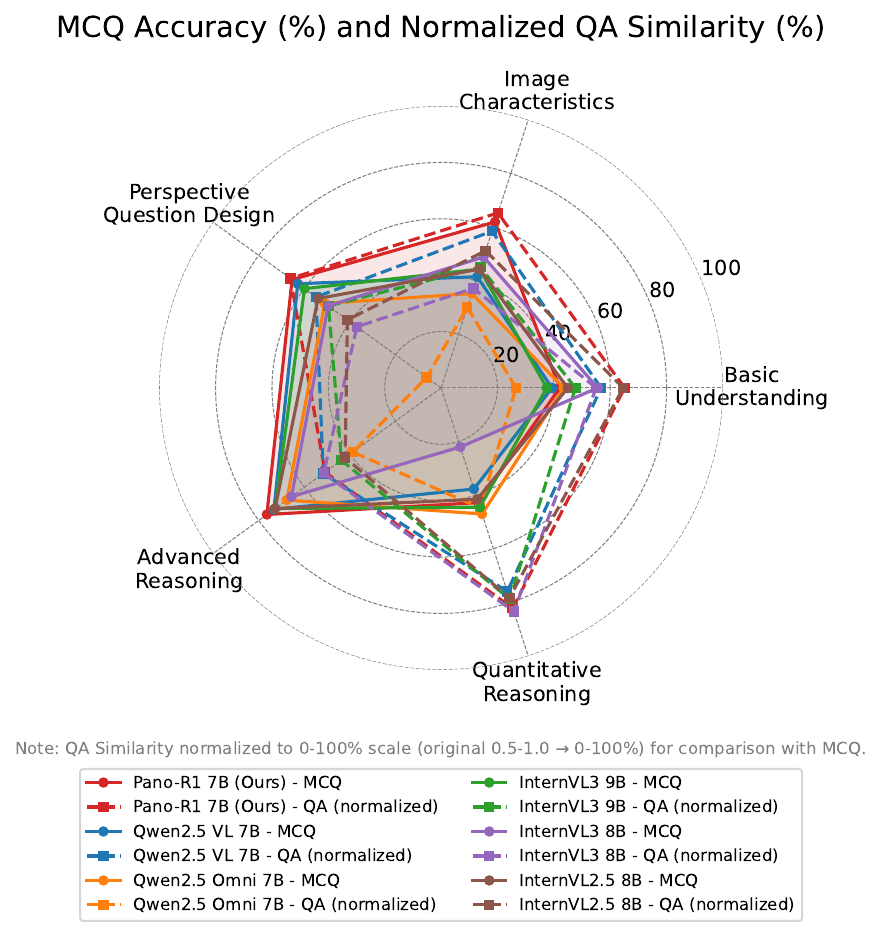}  
\caption{The radar chart comparing MCQ Accuracy (\%) and normalized QA Similarity (\%) across models and categories. Solid lines represent MCQ; dashed lines represent QA (normalized from 0.5-1.0 to 0-100\% for comparison).}
\Description{This is a combined polar radar chart that evaluates six vision-language models across five performance categories: Basic Understanding, Image Characteristics, Perspective Question Design, Advanced Reasoning, and Quantitative Reasoning. The chart plots two metrics for each model: MCQ Accuracy (solid lines with markers) and QA Similarity (dashed lines). To facilitate comparison, the QA scores, originally on a 0.5-1.0 scale, are normalized to the 0-100\% scale of the radial axis. Each model is assigned a unique color, with the area under its MCQ line lightly shaded. The chart visually highlights the performance of the 'Pano-R1 7B (Ours)' model, which generally demonstrates superior and more balanced results across all categories compared to the other five models.}
\label{fig:combined-radar}
\end{figure}

Omnidirectional scene understanding with 360$\circ$ cameras plays important roles in a wide range of applications, such as embodied AI, autonomous driving, and virtual/augmented reality (VR/AR)~\cite{Danieau2017EmbodimentOmniVideo,Li2021ARCampusNavigation,Kumar2021OmniDet}. Unlike conventional narrow-field vision, omnidirectional imagery captures the targeted scene with holistic Field-of-View (FoV) of 360$^\circ$ $\times$ 180$^\circ$, posing unique challenges such as severe geometric distortion, increased object density, and complex spatial relationships. These factors complicate visual perception and reasoning tasks especially spatial intelligence, making robust omnidirectional understanding an open research challenge~\cite{zhang2025towards,cao2025unlocking,zheng2025mllms,zhong2025omnisam,Zheng2023DualPathUDA,Zheng2023DistortionAwareUDA,Zheng2024SourceFreeUDA,Zheng2025SFUDAplusplus,cai2024interact360,dongfang2025multimodal}.

Multi-modal large language models (MLLMs) have spurred significant progress in visual reasoning tasks. However, most benchmarks and datasets are designed for standard perspective images, with only a few efforts—\textit{e.g.}, VQA360°~\cite{Chou2020VQA360}, Pano-AVQA~\cite{Yun2021PanoAVQA}, and OSR-Bench ~\cite{dongfang2025OSRbench}—beginning to address the demands of panoramic understanding. Nevertheless, these datasets typically lack comprehensive coverage of cross-frame, multi-view, and fine-grained reasoning—key aspects for real-world panoramic scene interpretation. As a result, the capabilities of modern MLLMs in performing complex reasoning across multi-view omnidirectional environments remain largely uncharacterized.

More recently, reasoning models, such as DeepSeek-R1, have demonstrated that reinforcement learning—particularly Group Relative Policy Optimization (GRPO)—can substantially enhance the capabilities of large models in coding, mathematics, and scientific reasoning~\cite{deepseekr1}. Building on these advancements, works such as Visual-RFT, VLM-R1, and R1-OneVision have successfully applied GRPO to large-scale MLLMs, yielding strong performance on tasks including visual STEM reasoning, referring expression comprehension, open-vocabulary object detection, and agentic multimodal tasks~\cite{liu2025visual,liu2025visualagenticreinforcementfinetuning,huang2025visionr1,shen2025vlmr1,yang2025r1}. A critical factor underlying these successes lies in the meticulous design of reward functions, which supply stable and interpretable learning signals for complex multimodal outputs. Motivated by these observations, we present Pano-R1 based on GRPO and design task-specific reward functions to ensure robust and consistent performance in panoramic cross-view VQA.

\begin{figure*}[ht]
  \centering
  \includegraphics[width=\textwidth]{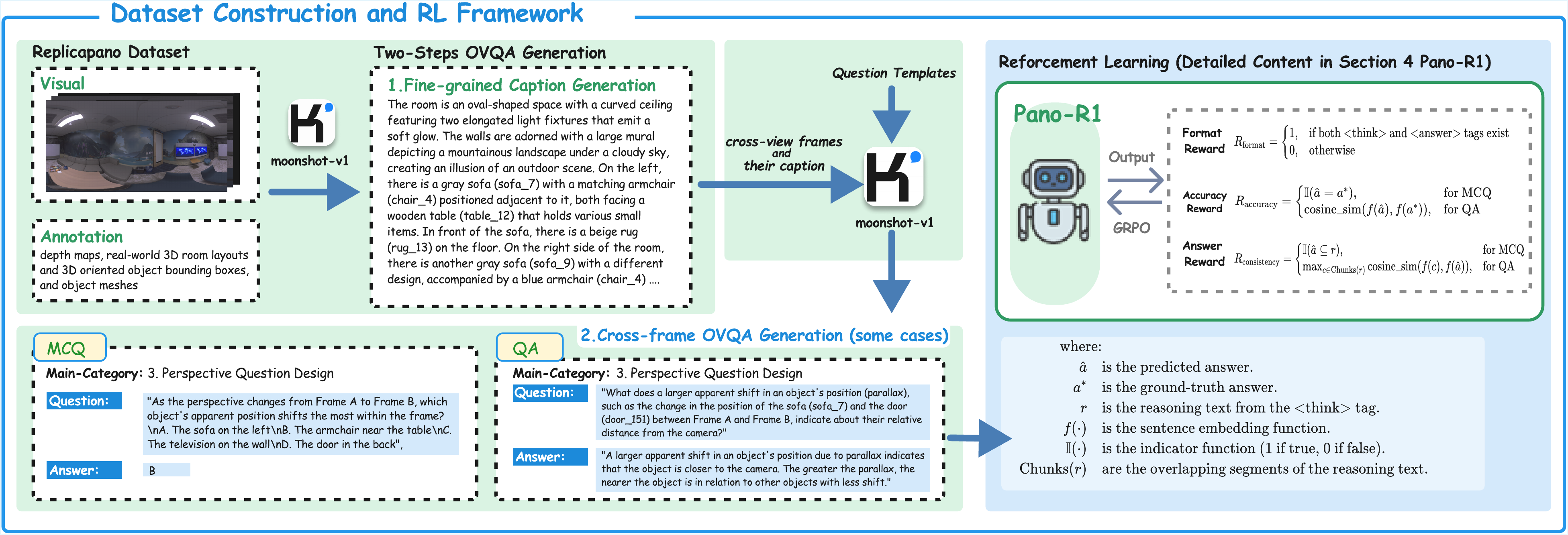}
  \caption{The overall framework for dataset construction and Reinforcement Learning training of the Pano-R1 model.}
  \Description{This diagram illustrates our end-to-end pipeline. The process begins with panoramic scenes from the Replicapano dataset. First, we employ a powerful vision-language model (moonshot-v1) to generate detailed, fine-grained captions for each scene. These captions, paired with structured question templates, are then used to systematically create the cross-frame Omnidirectional Visual Question Answering (OVQA) dataset, which includes both multiple-choice (MCQ) and open-ended question-answer (QA) pairs. Subsequently, our Pano-R1 model is trained on this generated data using a GRPO reinforcement learning strategy. The training is guided by a custom-designed, multi-component reward function that evaluates the output's format, accuracy, and the consistency between the reasoning and the final answer, ensuring the model learns to produce structured, accurate, and coherent responses.}
  \label{fig:framework}
\end{figure*}

To this end, we introduce \textbf{CFpano}, the first panoramic VQA dataset explicitly designed to evaluate cross-frame and cross-view understanding in 360° environments. By leveraging 3D annotations and fine-grained scene information from the base ReplicaPano dataset, CFpano enables systematic evaluation across a range of cognitive tasks, such as spatial relation understanding, occlusion reasoning, object identification, and quantitative reasoning. Our dataset comprises over 8,000 meticulously curated questions spanning a diverse array of reasoning dimensions.
Building upon our CFpano dataset, we introduce \methodname, a novel MLLM fine-tuned with GRPO and integrated with a set of task-specific reward functions. Our reward design includes format reward, accuracy reward, and consistency reward, which together guide the model toward generating well-structured and reliable answers. The complete framework is shown in Figure~\ref{fig:framework}.
% % 加一下reward function
% Pano-R1 is designed to excel in cross-frame panoramic reasoning, incorporating reward functions for format adherence, accuracy, and reasoning consistency during post-training.

Extensive experiments demonstrate that Pano-R1 achieves state-of-the-art performance on CFpano, consistently surpassing existing models across both multiple-choice and open-ended question-answering tasks. Our ablation studies further validate the efficacy of our reinforcement learning pipeline and reward design.
In summary, our key contributions are:
(1) We introduce \textbf{CFpano}, the first dataset dedicated to cross-frame panoramic visual question answering, which enables comprehensive evaluation of MLLMs in 360° environments.
(2) We propose \methodname, a reinforcement learning-enhanced MLLM tailored for panoramic scene understanding and reasoning.
(3) We conduct extensive experiments, establishing new baselines and offering in-depth analyses of model performance and training strategies on this challenging task.

\section{Related Work}

\subsection{Omnidirectional Understanding}

Omnidirectional understanding presents distinct challenges compared to conventional narrow-field vision. The widely-applied equirectangular projection in 360° imagery introduces inevitable geometric distortion, while the expanded field of view increases object density and spatial complexity, complicating scene interpretation. Recent efforts address these challenges through constructing datasets for omnidirectional visual question answering. VQA 360°~\cite{Chou2020VQA360} pioneers this direction, followed by Pano-AVQA~\cite{Yun2021PanoAVQA} which extends evaluation to 360° videos with spherical spatial and cross-modal reasoning tasks. OmniVQA~\cite{zhang2025360r1} targets object identification, attribute analysis, and spatial reasoning under polar distortion zone. OSR-Bench~\cite{dongfang2025OSRbench} systematically assesses panoramic spatial reasoning through negative sampling for hallucination robustness and a two-stage protocol involving omni-cognitive map reconstruction. Despite these advancements, existing datasets exhibit limitations in evaluating MLLMs under multi-frame cross-view scenarios. To bridge this gap, we introduce \textbf{CFpano}, the first dataset specifically designed for scene understanding and reasoning in 360° environments, enabling comprehensive MLLM evaluation.

\subsection{Multi-Modal Large Language Models}

MLLMs demonstrate significant advancements in visual tasks~\cite{lyu2024omnibind, Lyu2024UniBind, pmlr-BLIP, Li2023BLIP2, ACL024-revolution, Das2017VisualDialog, pmlr-xuc15}. Qwen2.5-VL~\cite{bai2025qwen25vl} enhances vision-language capabilities through dynamic resolution processing and absolute time encoding, enabling robust object localization, document parsing, and long-duration video comprehension. Concurrently, Qwen2.5-Omni~\cite{xu2025qwen25omni}---an end-to-end MLLM for unified perception across text, images, audio, and video---employs a novel Thinker-Talker architecture and Time-aligned Multimodal RoPE (TMRoPE) to minimize inter-modal interference while enhancing fusion. InternVL 2.5~\cite{chen2024internvl2.5} pioneered dynamic high-resolution training with progressive scaling strategies, optimizing vision-language alignment via multi-stage training. Its successor, InternVL 3~\cite{zhu2025internvl3}, revolutionizes training through native multimodal pre-training that unifies language and multimodal learning in a single stage, eliminating alignment bottlenecks. This approach achieves state-of-the-art performance on MMMU~\cite{yue2023mmmu} and long-video understanding benchmarks~\cite{wu2024longvideobench}. 
In this paper, we extend Qwen2.5-VL to the omnidirectional space and introduce \methodname, which incorporates rule-based reinforcement learning during post-training.
% In this paper, we propose \methodname, which incorporates rule-based reinforcement learning during post-training.

\subsection{Reinforcement Learning with Large Models}

Reinforcement learning (RL) techniques enhance large language model (LLM) performance through methods including Reinforcement Learning from Human Feedback (RLHF)~\cite{ouyang2022RLHF}, Proximal Policy Optimization (PPO)~\cite{schulman2017ppo}, Direct Preference Optimization (DPO)~\cite{rafailov2023dpo}, and Group Relative Policy Optimization (GRPO)~\cite{shao2024deepseekmath}. Recent extensions to large vision-language models (LVLMs) demonstrate promising results: Visual-RFT~\cite{liu2025visual}, VLM-r1~\cite{shen2025vlmr1}, and R1-OneVision~\cite{yang2025r1} exhibit strong reasoning capabilities with notable benchmark performance. In this work, we adapt GRPO for post-training and design three specialized reward functions for our proposed cross-frame omnidirectional VQA task.

% \begin{figure}[t!]  % [htbp] indicates floating position: here, top, bottom, page
%     \includegraphics[width=0.48\textwidth]{Figures/Dataset_cropped.pdf}
%     \caption{{Overview of the question distribution in the CFPano VQA dataset.} 
%     \Description{} 
%     \label{fig:dataset}  
% \end{figure}

\begin{figure}[ht]
\centering
\includegraphics[width=0.99\columnwidth]{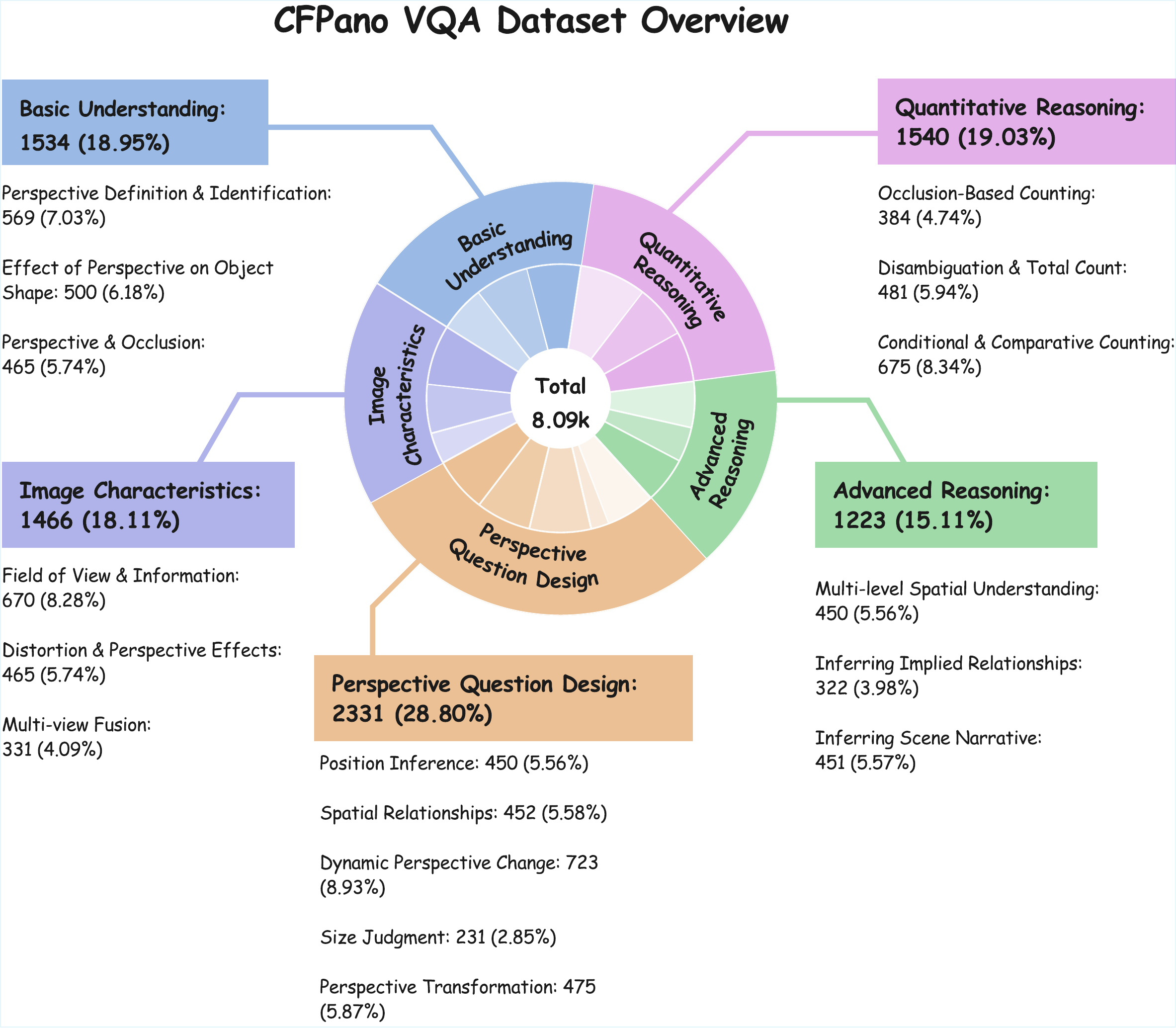}  
\caption{Hierarchical structure and question distribution of the CFPano VQA dataset.}
\Description{This figure provides a visual and statistical breakdown of the CFPano VQA dataset. A central donut chart shows the distribution of 8,094 questions across five main categories: Perspective Question Design (28.80\%), Quantitative Reasoning (19.03\%), Basic Understanding (18.95\%), Image Characteristics (18.11\%), and Advanced Reasoning (15.11\%). Each main category, represented by a colored segment, is connected to a corresponding box that further subdivides it into specific sub-tasks, detailing the exact number and percentage of questions for each. This hierarchical organization illustrates the dataset's comprehensive coverage for evaluating a wide range of visual reasoning skills.}
\label{fig:dataset}
\end{figure}

% \begin{figure*}
%   \includegraphics[width=\textwidth]{Figures/framewotk_cropped.pdf}
%   \caption{We first generate fine-grained captions for each panoramic scene with moonshot-v1-8k-vision-preview. These captions, combined with carefully designed question templates, enable systematic generation of cross-frame Omnidirectional Visual Question Answering (OVQA). Finally, Pano-R1 model is trained using GRPO with tailored reward functions, ensuring structured and consistent answers. }
%   \label{fig:framework}
% \end{figure*}

\section{The CFpano Dataset}

To evaluate the scene understanding and reasoning abilities of MLLMs in the cross-frame panoramic settings, we introduce the CFpano dataset, the first panoramic VQA benchmark that specifically focus on correlated panorama reasoning. CFpano is constructed based on the ReplicaPano dataset~\cite{Dong2024PanoContextFormer}, which provides diverse ground-truth annotations, including panoramic images, depth maps, real-world 3D room layouts, 3D-oriented object bounding boxes, and object meshes. These 3D annotations, together with object boundaries and meshes, greatly facilitate the generation of fine-grained captions and the design of VQA tasks and also ensure the correctness of our constructed ground truth data.The dataset overview is shown in Figure\ref{fig:dataset}.

\subsection{Overview}

Our  dataset  consists of \textbf{8,094} questions, with 64.74\% formatted as multiple-choice and 35.26\% as open-ended QA. The questions are systematically categorized as: the majority focus on perspective question design (28.80\%), followed by quantitative reasoning (19.03\%), basic understanding (18.95\%), image characteristics (18.11\%), and advanced reasoning (15.11\%). At a more granular level, the largest sub-categories are dynamic perspective change (8.93\%), conditional and comparative counting (8.34\%), field of view and information (8.28\%), and perspective definition and identification (7.03\%). The dataset is split into training and testing sets, maintaining a consistent distribution of question types and categories. This ensures comprehensive coverage of visual reasoning tasks across various levels of complexity and cognitive demand. 

\subsection{Question Template Design}

To systematically assess MLLMs' understanding and reasoning across cross-frame panoramic scenes, we design a structured template system for generating VQA questions. The templates cover multiple cognitive tasks, including basic perception, spatial relations, viewpoint changes, and occlusion reasoning. Each question template is organized by main category, sub-category, template ID, and textual content, supporting automated large-scale question generation.
Each template uses placeholders (such as \texttt{\{frame\_X\}} or \texttt{\{object\_A\}}), which are dynamically replaced with specific frame numbers or object names by leveraging the scene's 3D structure and semantic information during data generation. This template-based approach ensures both consistency and diversity in question generation. It enhances the controllability and comprehensiveness of the questions, enabling fine-grained analysis of model performance across different reasoning dimensions. This provides a solid data foundation for evaluating and advancing MLLMs.

\subsection{Cross-frame OVQA Generation}
Our VQA generation follows a two-stage process. First, using 3D information, object boundaries, and object meshes, we employ the Kimi vision model (\texttt{moonshot-v1-8k-vision-preview}) to generate fine-grained captions for each image. Second, for each scene, we randomly sample image pairs whose IDs differ by no more than 20, ensuring that the selected images share some visual content but do not differ significantly in viewpoint. From the carefully designed question templates (more details of the templates can be found in Table~\ref{tab:question_templates} in appendix), we randomly choose a template, and then use the Kimi vision model to generate answers based on both the question and the fine-grained captions. Finally, human annotators perform random sampling checks to ensure the overall data quality.

\section{Pano-R1}

To enhance understanding and reasoning capabilities in cross-frame omnidirectional environments, we propose Pano-R1. It is trained using Group Relative Policy Optimization (GRPO) with well-designed reward functions.

\subsection{Group Relative Policy Optimization}

Unlike reinforcement learning algorithms such as PPO, which require an additional critic model to estimate policy performance, GRPO directly compares groups of candidate responses, eliminating the need for a separate critic. This approach encourages the model to learn effective strategies by leveraging group context.
Given a question $q$, GRPO samples $N$ candidate responses $\{o_1, o_2, ..., o_N\}$ from the policy $\pi_\theta$ and evaluates each response $o_i$ using a reward function $R(q, o_i)$, which measures the quality of the candidate in the context of the given question. To determine the relative quality of these responses, GRPO normalizes the rewards by computing their mean and standard deviation and subsequently derives the advantage as:
\begin{equation}
A_i = \frac{r_i - \bar{r}}{\sigma},
\end{equation}
where $r_i = R(q, o_i)$, $\bar{r} = \operatorname{mean}\{r_1, \dots, r_N\}$, and $\sigma = \operatorname{std}\{r_1, \dots, r_N\}$. GRPO encourages the model to generate responses with higher advantages within the group by updating the policy $\pi_\theta$ using
\begin{equation}
\begin{aligned}
J_{\text{GRPO}}(\theta) = \mathbb{E}_{\{o_i\}_{i=1}^N \sim \pi_{\theta_{\text{old}}}(q)} \Bigg[ &\frac{1}{N} \sum_{i=1}^N \min(s_1 \cdot A_i, s_2 \cdot A_i) \\
&- \beta D_{\text{KL}}[\pi_\theta || \pi_{\text{ref}}] \Bigg],
\end{aligned}
\end{equation}
where
\begin{equation}
s_1 = \frac{\pi_\theta(o_i | q)}{\pi_{\theta_{\text{old}}}(o_i | q)}, s_2 = \operatorname{clip}\left( \frac{\pi_\theta(o_i | q)}{\pi_{\theta_{\text{old}}}(o_i | q)}, 1 - \epsilon, 1 + \epsilon \right).
\end{equation}
% \begin{equation}
% s_2 = \operatorname{clip}\left( \frac{\pi_\theta(o_i | q)}{\pi_{\theta_{\text{old}}}(o_i | q)}, 1 - \epsilon, 1 + \epsilon \right).
% \end{equation}

\subsection{Reward Functions}

For the VQA task, we design a composite reward function that evaluates model outputs along three key dimensions: format correctness, answer accuracy, and reasoning consistency. These rewards are tailored for use in GRPO, providing a comprehensive measure of response quality.

\noindent \textbf{Format Reward}  
It ensures that the model adheres to a structured output format by parsing the content within specific tags, \textit{i.e.}, \texttt{<think>...</think>} and \texttt{<answer>...</answer>}
% Specifically, it extracts text from \texttt{<think>...</think>} and \texttt{<answer>...</answer>}. 
If either tag is missing, the reward is zero; otherwise, it is one. This promotes consistent and parsable outputs. The reward function is:
\begin{equation}  
R_{\text{format}} =  
\begin{cases}  
1, & \text{if both tags exist} \\  
0, & \text{otherwise.}  
\end{cases}  
\end{equation}  
\noindent \textbf{Accuracy Reward}
The accuracy reward measures how well the predicted answer matches the ground truth, with tailored computations for multiple-choice questions (MCQ) and open-ended questions (QA). It encourages precise and relevant answers.  
For MCQ, the reward is binary, assigning 1 if the prediction exactly matches the ground truth and 0 otherwise.  
\begin{equation}  
R_{\text{answer}}^{\text{MCQ}} =  
\begin{cases}  
1, & \text{if prediction matches ground truth} \\  
0, & \text{otherwise.}  
\end{cases}  
\end{equation}  
For QA, the reward uses cosine similarity of embeddings to capture semantic closeness between the predicted answer and ground truth.  
\begin{equation}  
R_{\text{answer}}^{\text{QA}} = \operatorname{cosine\_sim}(f(\hat{a}), f(a^*)),  
\end{equation}  
where $\hat{a}$ is the predicted answer, $a^*$ is the ground-truth, $f(\cdot)$ is the embedding function from a pre-trained SentenceTransformer model, and $\operatorname{cosine\_sim}$ computes the cosine similarity (in $[0, 1]$).  

\noindent \textbf{Consistency Reward}
It promotes coherent reasoning by ensuring alignment between the reasoning process and the final answer. For MCQ, it verifies if the predicted answer option is mentioned in the reasoning text, providing a binary score.  
\begin{equation}  
R_{\text{consistency}} =  
\begin{cases}  
1, & \text{if MCQ and answer appears in reasoning} \\  
0, & \text{otherwise.}  
\end{cases}  
\end{equation}  
For QA, it computes the semantic similarity between the reasoning text (from \texttt{<think>}) and the answer (from \texttt{<answer>}) using embeddings. If the reasoning is long, it is divided into overlapping chunks, and the maximum similarity is taken. This ensures the reasoning logically supports the answer. 
\begin{equation}  
R_{\text{consistency}} = \max_{c \in \operatorname{Chunks}(r)} \operatorname{cosine\_sim}(f(c), f(\hat{a})),  
\end{equation} 
where $r$ is the reasoning text and $\operatorname{Chunks}(r)$ includes all overlapping segments and the full text.  

\paragraph{\textbf{Overall Reward}} It integrates the individual components to provide a balanced evaluation. It multiplies the format reward by the geometric mean of the clipped answer and consistency rewards, ensuring all aspects are essential for a high score. This formulation is crucial for robust multi-frame VQA reasoning.  

\begin{equation}  
\operatorname{Reward} = R_{\text{format}} \cdot \sqrt{ \operatorname{clip}(R_{\text{answer}}, 0, 1) \times \operatorname{clip}(R_{\text{consistency}}, 0, 1) },  
\end{equation}  

where $\operatorname{clip}(x, 0, 1)$ clamps $x$ to the interval $[0, 1]$, and $R_{\text{answer}}$ refers to either $R_{\text{answer}}^{\text{MCQ}}$ or $R_{\text{answer}}^{\text{QA}}$ depending on the question type.

\section{Experiments}

\begin{table*}[ht]
\centering
\small
\caption{Comparison of model performance across different reasoning categories on CFpano.}
\label{tab:model_performance}
\resizebox{\textwidth}{!}{
\begin{tabular}{lcc|cc|cc|cc|cc|cc}
\toprule
\multirow{3}{*}{Model} & \multicolumn{2}{c|}{Overall} & \multicolumn{2}{c|}{1. Basic} & \multicolumn{2}{c|}{2. Image} & \multicolumn{2}{c|}{3. Perspective} & \multicolumn{2}{c|}{4. Advanced} & \multicolumn{2}{c}{5. Quantitative} \\
& \multicolumn{2}{c|}{Performance} & \multicolumn{2}{c|}{Understanding} & \multicolumn{2}{c|}{Characteristics} & \multicolumn{2}{c|}{Question Design} & \multicolumn{2}{c|}{Reasoning} & \multicolumn{2}{c}{Reasoning} \\
& MCQ & QA & MCQ & QA & MCQ & QA & MCQ & QA & MCQ & QA & MCQ & QA \\
\midrule
\multicolumn{13}{l}{\textit{Qwen2.5 VL Series}} \\
Qwen2.5 VL 7B & 50.44\% & 0.7910 & 39.35\% & 0.7838 & 41.48\% & 0.7923 & \underline{62.87\%} & 0.7743 & \underline{73.33\%} & 0.7583 & 37.71\% & 0.8798 \\
Qwen2.5 VL 3B & 46.34\% & \underline{0.7967} & 40.74\% & 0.7975 & \underline{55.68\%} & \underline{0.7945} & 47.52\% & \underline{0.7895} & 55.38\% & \underline{0.7634} & 36.02\% & 0.8512 \\
\midrule
\multicolumn{13}{l}{\textit{Qwen2.5 Omni Series}} \\
Qwen2.5 Omni 7B & 48.98\% & 0.6092 & 43.52\% & 0.6329 & 35.23\% & 0.6503 & 50.99\% & 0.5316 & 67.69\% & 0.6936 & \textbf{47.03\%} & 0.7238 \\
Qwen2.5 Omni 3B & 43.85\% & 0.6150 & 38.89\% & 0.6296 & 30.11\% & 0.6063 & 43.56\% & 0.5537 & 72.31\% & 0.6627 & 35.17\% & 0.7950 \\
\midrule
\multicolumn{13}{l}{\textit{InternVL3 Series}} \\
InternVL3 9B & \underline{51.41}\% & 0.7571 & 37.50\% & 0.7383 & 44.32\% & 0.7254 & 59.90\% & 0.7470 & 72.82\% & 0.7184 & \underline{44.49\%} & 0.8940 \\
InternVL3 8B & 47.41\% & 0.7319 & \textbf{55.56\%} & 0.7750 & 48.86\% & 0.6858 & 49.50\% & 0.6846 & 65.64\% & 0.7574 & 22.03\% & \textbf{0.9173} \\
InternVL3 2B & 41.95\% & 0.6693 & \underline{52.31\%} & 0.6978 & 42.05\% & 0.6793 & 38.12\% & 0.6303 & 56.41\% & 0.7416 & 23.73\% & 0.7070 \\
\midrule
\multicolumn{13}{l}{\textit{InternVL2.5 Series}} \\
InternVL2.5 8B & 51.12\% & 0.7538 & 44.91\% & \underline{0.8230} & 44.32\% & 0.7558 & 53.96\% & 0.7055 & 72.82\% & 0.7103 & 41.53\% & 0.8918 \\
InternVL2.5 4B & 46.93\% & 0.7748 & 49.54\% & 0.7942 & 44.89\% & 0.7650 & 47.03\% & 0.7411 & 70.26\% & \textbf{0.7799} & 26.69\% & 0.8882 \\
InternVL2.5 2B & 42.05\% & 0.6509 & 44.44\% & 0.7536 & 42.61\% & 0.6519 & 40.10\% & 0.5919 & 64.10\% & 0.7002 & 22.88\% & 0.7249 \\
\midrule
\multicolumn{13}{l}{\textit{Our Model}} \\
\methodname (7B) & \textbf{56.78\%} & \textbf{0.8316} & 42.13\% & \textbf{0.8255} & \textbf{61.93\%} & \textbf{0.8264} & \textbf{65.35\%} & \textbf{0.8304} & \textbf{76.41\%} & 0.7545 & 42.80\% & \underline{0.9110} \\
\bottomrule
\end{tabular}
}
\end{table*}

\subsection{Experimental Setup}

We adopt the Qwen2.5-VL series (3B and 7B versions) as our base models, fine-tuning them on over 6k trainset of the CFPano dataset. For parameter-efficient tuning, both Supervised Fine-Tuning (SFT) and our Reinforcement Learning (RL) experiments with the GRPO algorithm employ LoRA with a rank of 64 and alpha of 128. All models are trained for 2 epochs with a learning rate of $1 \times 10^{-5}$. To optimize training, we freeze the vision encoder and utilize bfloat16 mixed-precision, Flash Attention 2, and gradient checkpointing on 4 A800 GPUs.

Our evaluation framework assesses both multiple-choice (MCQ) and open-ended question-answering (QA) performance. For MCQ, we measure exact match accuracy of the selected option. For QA, we compute semantic similarity using Sentence-BERT (all-MiniLM-L6-v2). To robustly handle varied model outputs, answers are extracted using a hierarchical strategy: first by parsing XML-style `<answer>` tags, with a fallback to a DeepSeek API-based method when necessary. This ensures reliable scoring across different response formats. 
%Further details can be found in the supplementary material.
Further details regarding the dataset construction, training process, and benchmarks can be found in the appendices.

\subsection{Main results}
Table~\ref{tab:model_performance} presents a comprehensive quantitative performance comparison of Pano-R1 against a suite of contemporary models, including the Qwen2.5 VL, Qwen2.5 Omni, InternVL3, and InternVL2.5 series. The evaluation is conducted across six key areas: Overall Performance, Basic Understanding, Image Characteristics, Perspective Question Design, Advanced Reasoning, and Quantitative Reasoning, with metrics provided for both Multiple-Choice Questions (MCQ) and direct Question Answering (QA). Pano-R1 demonstrates superior performance across the board, achieving the highest overall scores. Specifically, Pano-R1 obtains an Overall Performance MCQ score of 56.78\% and a QA score of 0.8316. This marks a significant improvement over the next best-performing models, InternVL3 9B (51.41\% MCQ) and Qwen2.5 VL 3B (0.7967 QA).

In the detailed sub-task analysis, Pano-R1 consistently ranks as a top performer. It achieves the highest scores in both MCQ and QA for Image Characteristics (61.93\%, 0.8264) and Perspective Question Design (65.35\%, 0.8304). Furthermore, it leads in Advanced Reasoning MCQ with a score of 76.41\%. While not leading in every single metric, such as Basic Understanding MCQ and Quantitative Reasoning MCQ, Pano-R1 maintains highly competitive performance, often securing the highest or near-highest QA scores in those categories. For instance, in Basic Understanding, it achieves the top QA score of 0.8255, and its Quantitative Reasoning QA score of 0.9110 is highly competitive.

\begin{figure*}[t!]
\centering
\includegraphics[width=0.99\textwidth]{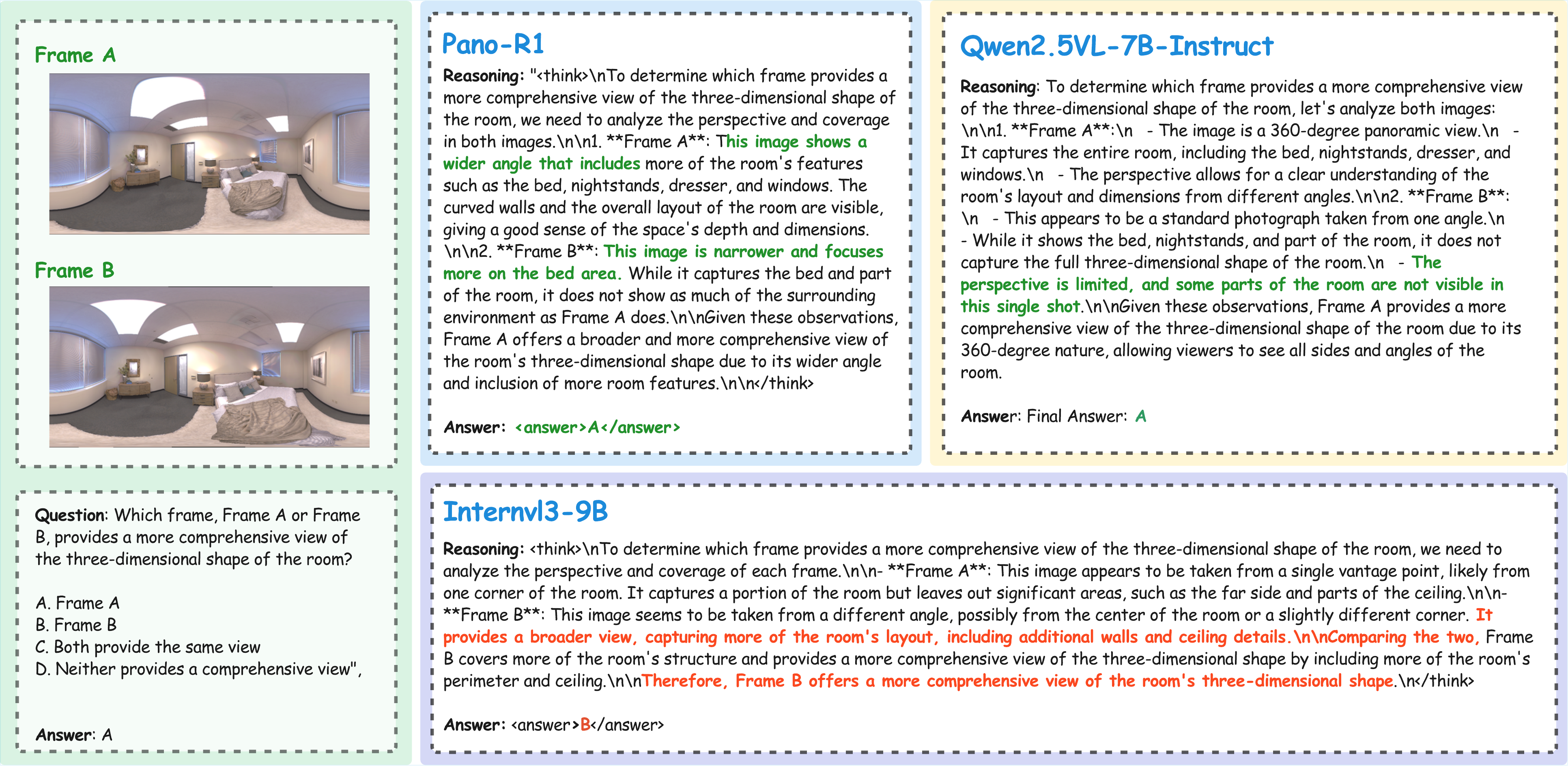}
\caption{The first case of model inference between our model (Pano-R1) and other leading MLLMs.}
\Description{This figure provides a side-by-side comparison of how three different models—Pano-R1, Qwen2.5VL-7B-Instruct, and InternVL3-9B—respond to a visual reasoning question. The task is to identify which of two images (Frame A, a panoramic view, and Frame B, a standard view) offers a more comprehensive perspective of a room. The outputs show that Pano-R1 correctly identifies Frame A, provides sound reasoning, and adheres to the required structured output format with `<think>` and `<answer>` tags. Qwen2.5VL-7B also reasons correctly and provides the right answer but fails to follow the specified format. InternVL3-9B fails on both fronts, presenting flawed reasoning and an incorrect answer. This case study highlights Pano-R1's effectiveness in both complex reasoning and instruction following.}
\label{fig:inference-case3}
\end{figure*}

This figure \ref{fig:inference-case3} presents a comparison of three models—Pano-R1, Qwen2.5VL-7B-Instruct, and InternVL3-9B—on a visual reasoning task involving panoramic images. Two example images (Frame A and Frame B) of a room are shown. Each model’s predicted answer and reasoning are displayed in colored boxes. Pano-R1 and Qwen2.5VL-7B-Instruct both select the correct answer (“A”), but only Pano-R1 produces output in the required answer format. InternVL3-9B selects the wrong answer (“B”) and provides flawed reasoning. The figure highlights that Pano-R1’s output is not only accurate but also better aligned with the expected answer structure and reasoning depth, demonstrating the model’s robustness in both answer correctness and response formatting.

These results underscore the effectiveness of our model's architecture and training methodology, establishing a new state-of-the-art on this benchmark across a diverse set of visual reasoning tasks.

% \begin{figure*}[t!]
% \centering
% \includegraphics[width=0.98\textwidth]{Figures/Case3_cropped .pdf}
% \caption{Qualitative results comparison: Qwen2.5VL-7B-Instruct's reasoning and answer are both correct but failed to output in the specified format. InternVL3-9B's reasoning and answer are both wrong. Pano-R1's reasoning and answer are both correct.}
% \label{fig:inference-case3}
% \end{figure*}

\subsection{Ablation Study}
\subsubsection{Analysis of Training Strategy}
To validate the effectiveness of our ptraining pipeline, we compare a model trained with only SFT against our final Pano-R1 model, which incorporates an additional RL phase. The results in Table~\ref{tab:train_method_compare} demonstrate that the inclusion of RL yields significant performance gains. Overall, the RL-enhanced model improves the MCQ score from 52.6\% to \textbf{56.8\%} and the QA score from 0.81 to \textbf{0.83}.
This trend is particularly evident in categories requiring deeper comprehension, such as Basic Understanding and Perspective Question Design, where MCQ scores see substantial increases. Furthermore, the QA scores are consistently improved or maintained across nearly all categories, highlighting the model's enhanced ability to generate accurate answers. While the SFT-only model achieves higher MCQ scores in Advanced and Quantitative Reasoning, our RL-trained model shows superior or equal QA performance in those same areas. This suggests that RL refines the model's response generation for greater factual accuracy, demonstrating its crucial role in achieving a more robust and capable model.

\begin{table}[ht]
\centering
\footnotesize % 换成 \small 以改善可读性（如果 \footnotesize 太小）
\caption{Training strategy comparison: SFT vs. RL (Pano-R1).}
\label{tab:train_method_compare}
\begin{tabular}{@{}l|cc|cc@{}}
\toprule % booktabs: 顶部粗线
\multirow{2}{*}{\textbf{Category}} & \multicolumn{2}{c|}{\textbf{Qwen2.5VL-7B + SFT}} & \multicolumn{2}{c@{}}{\textbf{Qwen2.5VL-7B + RL}} \\
 & MCQ & QA & MCQ & QA \\
\midrule % booktabs: 中间分隔线
Overall & 52.6\% & 0.81 & \textbf{56.8\%} & \textbf{0.83} \\
\midrule
Basic Understanding & 33.8\% & 0.81 & \textbf{42.1\%} & \textbf{0.83} \\
Image Characteristics & 50.6\% & \textbf{0.83} & \textbf{61.9\%} & 0.83 \\
Perspective Question Design & 55.5\% & 0.79 & \textbf{65.4\%} & \textbf{0.83} \\
Advanced Reasoning & \textbf{81.5\%} & 0.76 & 76.4\% & \textbf{0.76} \\
Quantitative Reasoning & \textbf{44.9\%} & 0.89 & 42.8\% & \textbf{0.91} \\
\bottomrule % booktabs: 底部粗线
\end{tabular}
\end{table}

\subsubsection{Ablation Study on Model Parameters Variants}
We further investigate the impact of model scale by comparing two variants: Pano-R1-3B and Pano-R1-7B. Both models are developed using the same training strategy, with the results of this comparison detailed in Table~\ref{tab:train_param_compare}. The analysis reveals that the larger Pano-R1-7B model holds a substantial advantage in the MCQ format, achieving a much higher overall score of \textbf{56.8\%} compared to 49.7\% for the 3B variant. This superiority is consistent across most categories, especially in complex reasoning tasks.
Conversely, the performance on the direct Question Answering (QA) metric is more nuanced. The overall QA scores are much closer, with the 7B model achieving \textbf{0.83} and the 3B model reaching 0.80. The larger 7B model establishes a clear lead in QA performance for categories like Basic, Image, and Quantitative Reasoning. This ablation highlights that while increased model capacity is critical for improving performance on selection-based tasks (MCQ), the benefits for direct answer generation (QA) are more varied across different reasoning types.

\begin{table}[ht]
\centering
\small % 建议使用 \small 以确保可读性
\caption{Performance comparison of different model sizes.}
\label{tab:train_param_compare}
\resizebox{\columnwidth}{!}{ % 添加缩放以适应宽度（可选，如果表格太宽）
\begin{tabular}{@{}l|cc|cc@{}}
\toprule % booktabs: 顶部粗线
\multirow{2}{*}{\textbf{Category}} & \multicolumn{2}{c|}{\textbf{Pano-R1-3B}} & \multicolumn{2}{c@{}}{\textbf{Pano-R1-7B}} \\
 & MCQ & QA & MCQ & QA \\
\midrule % booktabs: 中间分隔线
Overall & 49.7\% & 0.80 & \textbf{56.8\%} & \textbf{0.83} \\
\midrule
Basic Understanding & \textbf{44.9\%} & 0.79 & 42.1\% & \textbf{0.83} \\
Image Characteristics & 55.1\% & 0.80 & \textbf{61.9\%} & \textbf{0.83} \\
Perspective Question Design & 48.5\% & 0.81 & \textbf{65.4\%} & \textbf{0.83} \\
Advanced Reasoning & 64.1\% & 0.75 & \textbf{76.4\%} & \textbf{0.75} \\
Quantitative Reasoning & 39.0\% & 0.86 & \textbf{42.8\%} & \textbf{0.91} \\
\bottomrule % booktabs: 底部粗线
\end{tabular}
} % 结束 resizebox
\end{table}

\section{Conclusion}
In this paper, we addressed the challenges of visual question answering with cross-frame correlated panoramas by building the first CFpano, a comprehensive dataset designed for cross-frame and cross-view understanding in 360° environments. We further proposed Pano-R1, a MLLM enhanced with GRPO and designed task-specified reward functions to ensure reliable and structured reasoning. Extensive experiments demonstrated that Pano-R1 establishes new state-of-the-art results across diverse reasoning categories, highlighting the importance of reinforcement learning and reward design for robust omnidirectional scene understanding. We hope our dataset, model, and analyses will inspire future research on panoramic understanding and the development of more capable MLLMs for complex real-world environments.

\noindent \textbf{Future Works:} While Pano-R1 demonstrates strong performance in cross-frame panoramic VQA, several avenues remain for future exploration. First, expanding the CFpano dataset to include dynamic scenes, such as video sequences or outdoor environments, could enable evaluation of temporal reasoning in omnidirectional settings. Additionally, integrating advanced multimodal inputs, like depth maps or audio cues, may further enhance model robustness. Additionally, developing more efficient reinforcement learning algorithms tailored for panoramic environments are promising directions.

% \clearpage
\bibliographystyle{ACM-Reference-Format}
\bibliography{sample-base}

% \clearpage
\appendix

\section{Prompt Templates}
\subsection{Prompts for VQA generation}
\subsubsection{Fixed System Instruction (Base Part)}
\begin{tcolorbox}[colback=white, colframe=black, boxrule=0.3pt, arc=1pt, breakable, left=1mm, right=1mm, top=0.5mm, bottom=0.5mm]
\footnotesize\ttfamily
You are an expert in Visual Question Answering (VQA) generation. Your task is to create a question-answer pair based on two input images and their text descriptions.

**Rules:**
1. **Analyze Inputs**: Carefully analyze the two provided frames, identified as **Frame A** (original ID: \{frame\_x\_id\}) and **Frame B** (original ID: \{frame\_y\_id\}), and their corresponding captions.
2. **Use Template**: Generate a question that follows the structure of the provided template. You must use the placeholders "Frame A" and "Frame B" in your question, NOT the original IDs. You must also fill in other placeholders (like \{object\_A\}) with relevant objects or concepts visible in the images.
\end{tcolorbox}

\subsubsection{MCQ Format Instruction}
\begin{tcolorbox}[colback=white, colframe=black, boxrule=0.3pt, arc=1pt, breakable, left=1mm, right=1mm, top=0.5mm, bottom=0.5mm]
\footnotesize\ttfamily
3. **Format**: This is a Multiple-Choice Question (MCQ).
4. **JSON Output**: Your final output must be a single, valid JSON object with the following keys:
   - "question": The generated question string.
   - "options": A JSON object containing four plausible options, with keys "A", "B", "C", and "D". One option must be the correct answer.
   - "answer": The string value of the correct option (e.g., the text from option "C").
   Do not add any text outside the JSON object.
\end{tcolorbox}

\subsubsection{QA Format Instruction}
\begin{tcolorbox}[colback=white, colframe=black, boxrule=0.3pt, arc=1pt, breakable, left=1mm, right=1mm, top=0.5mm, bottom=0.5mm]
\footnotesize\ttfamily
3. **Format**: This is an open-ended Question-Answer (QA).
4. **JSON Output**: Your final output must be a single, valid JSON object with the following keys:
   - "question": The generated question string.
   - "answer": A concise, factual, open-ended answer string.
   The JSON object for a QA must not contain an "options" key. Do not add any text outside the JSON object.
\end{tcolorbox}

\subsection{Prompts for fine-grained Caption generation}
\subsubsection{Image Description Generation Instruction}
\begin{tcolorbox}[colback=white, colframe=black, boxrule=0.3pt, arc=1pt, breakable, left=1mm, right=1mm, top=0.5mm, bottom=0.5mm]
\footnotesize\ttfamily
You are a meticulous and objective scene captioner. Your task is to generate a precise, factual, and comprehensive description of the provided 360-degree panoramic image and its metadata.

Your description must adhere to the following strict rules:
1. **Strictly Objective Tone**: Your language must be neutral and descriptive. DO NOT use interpretive, subjective, or emotional words. Avoid describing the "mood," "atmosphere," "feel," or "energy" (e.g., do not use words like "cozy," "inviting," "playful," "serene," or "elegant").
2. **Structural-First Approach**: Begin with the room's physical layout, architectural features (e.g., "The room is rectangular with white walls, a hardwood floor, and a flat white ceiling."), and light sources (e.g., "Natural light enters from a large window on the north wall.").
3. **Systematic Object Inventory**: Methodically describe each object from the metadata. For each object, state its key visual properties (color, material, shape, design) and its precise spatial relationship to other objects or room features.
4. **Factual and Comprehensive Report**: The final output must be a coherent, well-structured paragraph that reads like a technical report or an inventory list written in prose. It must not be a creative essay or a story. Ensure every object from the metadata is included.

Please generate the description now, following these rules precisely.
\end{tcolorbox}

\subsection{Prompts for result analysis}
\subsubsection{Answer Extraction Prompt of MCQ Type}
\begin{tcolorbox}[colback=white, colframe=black, boxrule=0.3pt, arc=1pt, breakable, left=1mm, right=1mm, top=0.5mm, bottom=0.5mm]
\footnotesize\ttfamily
You are a text parsing expert. Based on the following question and model output, your task is to extract only the final option letter (A, B, C, or D). Do not provide any explanation or reasoning. Return only the single letter.
Question Type: {question\_type}
Question: {question}
\end{tcolorbox}

\subsubsection{Answer Extraction Prompt of QA Type}
\begin{tcolorbox}[colback=white, colframe=black, boxrule=0.3pt, arc=1pt, breakable, left=1mm, right=1mm, top=0.5mm, bottom=0.5mm]
\footnotesize\ttfamily
You are a text parsing expert. Based on the following question and model output, your task is to extract only the final, direct answer. Do not include any of the model's thought process or introductory phrases like 'The answer is:'. Provide only the clean answer text.

Question Type: {question\_type}
Question: {question}
\end{tcolorbox}

\section{More Details of Model Performance on Dataset}
This appendix provides detailed performance metrics for 11 models on the dataset. Each model is evaluated across categories and sub-categories, with metrics for MCQ Accuracy and QA Similarity. Results are presented in separate small tables for clarity. 

\subsection{InternVL3-9B}
\begin{table}[ht]
\centering
\footnotesize
\caption{Performance Evaluation of InternVL3-9B}
\label{tab:internvl3-9b}
\resizebox{\columnwidth}{!}{
\begin{tabular}{p{5cm}cc}
\toprule
\textbf{Category/Sub-category} & \textbf{MCQ Accuracy} & \textbf{QA Similarity} \\
\midrule
\textbf{Overall Performance} & 51.41\% & 0.7571 \\
\midrule
\textbf{1. Basic Understanding} & 37.50\% & 0.7383 \\
1.1 Perspective Definition \& Identification & 29.63\% & 0.9089 \\
1.2 Effect of Perspective on Object Shape & 56.82\% & 0.8198 \\
1.3 Perspective \& Occlusion & 37.50\% & 0.5382 \\
\midrule
\textbf{2. Image Characteristics} & 44.32\% & 0.7254 \\
2.1 Field of View \& Information & 48.96\% & 0.6403 \\
2.2 Distortion \& Perspective Effects & 38.75\% & 0.8336 \\
2.3 Multi-view Fusion & -- & 0.7727 \\
\midrule
\textbf{3. Perspective Question Design} & 59.90\% & 0.7470 \\
3.1 Position Inference & 60.53\% & 0.7063 \\
3.2 Spatial Relationships & 64.44\% & 0.7356 \\
3.3 Dynamic Perspective Change & 63.49\% & 0.7868 \\
3.4 Size Judgment & 72.73\% & -- \\
3.5 Perspective Transformation & 21.74\% & 0.7334 \\
\midrule
\textbf{4. Advanced Reasoning} & 72.82\% & 0.7184 \\
4.1 Multi-level Spatial Understanding & 56.82\% & 0.6910 \\
4.2 Inferring Implied Relationships & 85.14\% & 0.9076 \\
4.3 Inferring Scene Narrative & 70.13\% & 0.7648 \\
\midrule
\textbf{5. Quantitative Reasoning} & 44.49\% & 0.8940 \\
5.1 Occlusion-Based Counting & 52.50\% & -- \\
5.2 Disambiguation \& Total Count & 38.10\% & 0.8779 \\
5.3 Conditional \& Comparative Counting & 41.94\% & 0.9041 \\
\bottomrule
\end{tabular}
}
\end{table}

\subsection{InternVL3-8B}
\begin{table}[H]
\centering
\footnotesize
\caption{Performance Evaluation of InternVL3-8B}
\label{tab:internvl3-8b}
\resizebox{\columnwidth}{!}{
\begin{tabular}{p{5cm}cc}
\toprule
\textbf{Category/Sub-category} & \textbf{MCQ Accuracy} & \textbf{QA Similarity} \\
\midrule
\textbf{Overall Performance} & 47.41\% & 0.7319 \\
\midrule
\textbf{1. Basic Understanding} & 55.56\% & 0.7750 \\
1.1 Perspective Definition \& Identification & 53.70\% & 0.4999 \\
1.2 Effect of Perspective on Object Shape & 50.00\% & 0.8197 \\
1.3 Perspective \& Occlusion & 62.50\% & 0.6839 \\
\midrule
\textbf{2. Image Characteristics} & 48.86\% & 0.6858 \\
2.1 Field of View \& Information & 61.46\% & 0.5985 \\
2.2 Distortion \& Perspective Effects & 33.75\% & 0.6816 \\
2.3 Multi-view Fusion & -- & 0.7420 \\
\midrule
\textbf{3. Perspective Question Design} & 49.50\% & 0.6846 \\
3.1 Position Inference & 63.16\% & 0.5401 \\
3.2 Spatial Relationships & 68.89\% & 0.8109 \\
3.3 Dynamic Perspective Change & 17.46\% & 0.7051 \\
3.4 Size Judgment & 69.70\% & -- \\
3.5 Perspective Transformation & 47.83\% & 0.6807 \\
\midrule
\textbf{4. Advanced Reasoning} & 65.64\% & 0.7574 \\
4.1 Multi-level Spatial Understanding & 77.27\% & 0.7402 \\
4.2 Inferring Implied Relationships & 55.41\% & 0.8915 \\
4.3 Inferring Scene Narrative & 68.83\% & 0.7857 \\
\midrule
\textbf{5. Quantitative Reasoning} & 22.03\% & 0.9173 \\
5.1 Occlusion-Based Counting & 8.75\% & -- \\
5.2 Disambiguation \& Total Count & 6.35\% & 0.8962 \\
5.3 Conditional \& Comparative Counting & 44.09\% & 0.9306 \\
\bottomrule
\end{tabular}
}
\end{table}

\subsection{InternVL3-2B}
\begin{table}[H]
\centering
\footnotesize
\caption{Performance Evaluation of InternVL3-2B}
\label{tab:internvl3-2b}
\resizebox{\columnwidth}{!}{
\begin{tabular}{p{5cm}cc}
\toprule
\textbf{Category/Sub-category} & \textbf{MCQ Accuracy} & \textbf{QA Similarity} \\
\midrule
\textbf{Overall Performance} & 41.95\% & 0.6693 \\
\midrule
\textbf{1. Basic Understanding} & 52.31\% & 0.6978 \\
1.1 Perspective Definition \& Identification & 56.48\% & 0.3506 \\
1.2 Effect of Perspective on Object Shape & 56.82\% & 0.7828 \\
1.3 Perspective \& Occlusion & 42.19\% & 0.5154 \\
\midrule
\textbf{2. Image Characteristics} & 42.05\% & 0.6793 \\
2.1 Field of View \& Information & 53.12\% & 0.6198 \\
2.2 Distortion \& Perspective Effects & 28.75\% & 0.5474 \\
2.3 Multi-view Fusion & -- & 0.7262 \\
\midrule
\textbf{3. Perspective Question Design} & 38.12\% & 0.6303 \\
3.1 Position Inference & 47.37\% & 0.5772 \\
3.2 Spatial Relationships & 57.78\% & 0.7974 \\
3.3 Dynamic Perspective Change & 20.63\% & 0.5080 \\
3.4 Size Judgment & 48.48\% & -- \\
3.5 Perspective Transformation & 17.39\% & 0.7153 \\
\midrule
\textbf{4. Advanced Reasoning} & 56.41\% & 0.7416 \\
4.1 Multi-level Spatial Understanding & 59.09\% & 0.7187 \\
4.2 Inferring Implied Relationships & 56.76\% & 0.8524 \\
4.3 Inferring Scene Narrative & 54.55\% & 0.7827 \\
\midrule
\textbf{5. Quantitative Reasoning} & 23.73\% & 0.7070 \\
5.1 Occlusion-Based Counting & 22.50\% & -- \\
5.2 Disambiguation \& Total Count & 3.17\% & 0.3876 \\
5.3 Conditional \& Comparative Counting & 38.71\% & 0.9084 \\
\bottomrule
\end{tabular}
}
\end{table}

\subsection{InternVL2.5-8B}
\begin{table}[H]
\centering
\footnotesize
\caption{Performance Evaluation of InternVL2.5-8B}
\label{tab:internvl2.5-8b}
\resizebox{\columnwidth}{!}{
\begin{tabular}{p{5cm}cc}
\toprule
\textbf{Category/Sub-category} & \textbf{MCQ Accuracy} & \textbf{QA Similarity} \\
\midrule
\textbf{Overall Performance} & 51.12\% & 0.7538 \\
\midrule
\textbf{1. Basic Understanding} & 44.91\% & 0.8230 \\
1.1 Perspective Definition \& Identification & 38.89\% & 0.5179 \\
1.2 Effect of Perspective on Object Shape & 59.09\% & 0.8528 \\
1.3 Perspective \& Occlusion & 45.31\% & 0.7682 \\
\midrule
\textbf{2. Image Characteristics} & 44.32\% & 0.7558 \\
2.1 Field of View \& Information & 38.54\% & 0.5961 \\
2.2 Distortion \& Perspective Effects & 51.25\% & 0.8493 \\
2.3 Multi-view Fusion & -- & 0.8518 \\
\midrule
\textbf{3. Perspective Question Design} & 53.96\% & 0.7055 \\
3.1 Position Inference & 63.16\% & 0.6672 \\
3.2 Spatial Relationships & 57.78\% & 0.6103 \\
3.3 Dynamic Perspective Change & 52.38\% & 0.7277 \\
3.4 Size Judgment & 51.52\% & -- \\
3.5 Perspective Transformation & 39.13\% & 0.7695 \\
\midrule
\textbf{4. Advanced Reasoning} & 72.82\% & 0.7103 \\
4.1 Multi-level Spatial Understanding & 50.00\% & 0.7001 \\
4.2 Inferring Implied Relationships & 87.84\% & 0.6823 \\
4.3 Inferring Scene Narrative & 71.43\% & 0.7326 \\
\midrule
\textbf{5. Quantitative Reasoning} & 41.53\% & 0.8918 \\
5.1 Occlusion-Based Counting & 43.75\% & -- \\
5.2 Disambiguation \& Total Count & 28.57\% & 0.8495 \\
5.3 Conditional \& Comparative Counting & 48.39\% & 0.9185 \\
\bottomrule
\end{tabular}
}
\end{table}

\subsection{InternVL2.5-4B}
\begin{table}[H]
\centering
\footnotesize
\caption{Performance Evaluation of InternVL2.5-4B}
\label{tab:internvl2.5-4b}
\resizebox{\columnwidth}{!}{
\begin{tabular}{p{5cm}cc}
\toprule
\textbf{Category/Sub-category} & \textbf{MCQ Accuracy} & \textbf{QA Similarity} \\
\midrule
\textbf{Overall Performance} & 46.93\% & 0.7748 \\
\midrule
\textbf{1. Basic Understanding} & 49.54\% & 0.7942 \\
1.1 Perspective Definition \& Identification & 55.56\% & 0.6769 \\
1.2 Effect of Perspective on Object Shape & 54.55\% & 0.8121 \\
1.3 Perspective \& Occlusion & 35.94\% & 0.7579 \\
\midrule
\textbf{2. Image Characteristics} & 44.89\% & 0.7650 \\
2.1 Field of View \& Information & 45.83\% & 0.6450 \\
2.2 Distortion \& Perspective Effects & 43.75\% & 0.7042 \\
2.3 Multi-view Fusion & -- & 0.8459 \\
\midrule
\textbf{3. Perspective Question Design} & 47.03\% & 0.7411 \\
3.1 Position Inference & 50.00\% & 0.7143 \\
3.2 Spatial Relationships & 55.56\% & 0.7867 \\
3.3 Dynamic Perspective Change & 41.27\% & 0.7116 \\
3.4 Size Judgment & 39.39\% & -- \\
3.5 Perspective Transformation & 52.17\% & 0.7687 \\
\midrule
\textbf{4. Advanced Reasoning} & 70.26\% & 0.7799 \\
4.1 Multi-level Spatial Understanding & 56.82\% & 0.7700 \\
4.2 Inferring Implied Relationships & 71.62\% & 0.8058 \\
4.3 Inferring Scene Narrative & 76.62\% & 0.7988 \\
\midrule
\textbf{5. Quantitative Reasoning} & 26.69\% & 0.8882 \\
5.1 Occlusion-Based Counting & 21.25\% & -- \\
5.2 Disambiguation \& Total Count & 25.40\% & 0.8485 \\
5.3 Conditional \& Comparative Counting & 32.26\% & 0.9132 \\
\bottomrule
\end{tabular}
}
\end{table}

\subsection{InternVL2.5-2B}
\begin{table}[H]
\centering
\footnotesize
\caption{Performance Evaluation of InternVL2.5-2B}
\label{tab:internvl2.5-2b}
\resizebox{\columnwidth}{!}{
\begin{tabular}{p{5cm}cc}
\toprule
\textbf{Category/Sub-category} & \textbf{MCQ Accuracy} & \textbf{QA Similarity} \\
\midrule
\textbf{Overall Performance} & 42.05\% & 0.6509 \\
\midrule
\textbf{1. Basic Understanding} & 44.44\% & 0.7536 \\
1.1 Perspective Definition \& Identification & 50.93\% & 0.8985 \\
1.2 Effect of Perspective on Object Shape & 45.45\% & 0.7982 \\
1.3 Perspective \& Occlusion & 32.81\% & 0.6417 \\
\midrule
\textbf{2. Image Characteristics} & 42.61\% & 0.6519 \\
2.1 Field of View \& Information & 60.42\% & 0.5565 \\
2.2 Distortion \& Perspective Effects & 21.25\% & 0.5092 \\
2.3 Multi-view Fusion & -- & 0.7225 \\
\midrule
\textbf{3. Perspective Question Design} & 40.10\% & 0.5919 \\
3.1 Position Inference & 28.95\% & 0.6784 \\
3.2 Spatial Relationships & 68.89\% & 0.7400 \\
3.3 Dynamic Perspective Change & 31.75\% & 0.4405 \\
3.4 Size Judgment & 45.45\% & -- \\
3.5 Perspective Transformation & 17.39\% & 0.6232 \\
\midrule
\textbf{4. Advanced Reasoning} & 64.10\% & 0.7002 \\
4.1 Multi-level Spatial Understanding & 61.36\% & 0.7019 \\
4.2 Inferring Implied Relationships & 54.05\% & 0.7924 \\
4.3 Inferring Scene Narrative & 75.32\% & 0.6920 \\
\midrule
\textbf{5. Quantitative Reasoning} & 22.88\% & 0.7249 \\
5.1 Occlusion-Based Counting & 10.00\% & -- \\
5.2 Disambiguation \& Total Count & 11.11\% & 0.5116 \\
5.3 Conditional \& Comparative Counting & 41.94\% & 0.8594 \\
\bottomrule
\end{tabular}
}
\end{table}

\subsection{Qwen2.5-Omni-7B}
\begin{table}[H]
\centering
\footnotesize
\caption{Performance Evaluation of Qwen2.5-Omni-7B}
\label{tab:qwen2.5-omni-7b}
\resizebox{\columnwidth}{!}{
\begin{tabular}{p{5cm}cc}
\toprule
\textbf{Category/Sub-category} & \textbf{MCQ Accuracy} & \textbf{QA Similarity} \\
\midrule
\textbf{Overall Performance} & 48.98\% & 0.6092 \\
\midrule
\textbf{1. Basic Understanding} & 43.52\% & 0.6329 \\
1.1 Perspective Definition \& Identification & 37.96\% & 0.4448 \\
1.2 Effect of Perspective on Object Shape & 54.55\% & 0.7747 \\
1.3 Perspective \& Occlusion & 45.31\% & 0.3091 \\
\midrule
\textbf{2. Image Characteristics} & 35.23\% & 0.6503 \\
2.1 Field of View \& Information & 39.58\% & 0.6211 \\
2.2 Distortion \& Perspective Effects & 30.00\% & 0.7135 \\
2.3 Multi-view Fusion & -- & 0.6648 \\
\midrule
\textbf{3. Perspective Question Design} & 50.99\% & 0.5316 \\
3.1 Position Inference & 44.74\% & 0.6156 \\
3.2 Spatial Relationships & 68.89\% & 0.5515 \\
3.3 Dynamic Perspective Change & 39.68\% & 0.4477 \\
3.4 Size Judgment & 54.55\% & -- \\
3.5 Perspective Transformation & 52.17\% & 0.5644 \\
\midrule
\textbf{4. Advanced Reasoning} & 67.69\% & 0.6936 \\
4.1 Multi-level Spatial Understanding & 63.64\% & 0.7152 \\
4.2 Inferring Implied Relationships & 77.03\% & 0.6633 \\
4.3 Inferring Scene Narrative & 61.04\% & 0.6510 \\
\midrule
\textbf{5. Quantitative Reasoning} & 47.03\% & 0.7238 \\
5.1 Occlusion-Based Counting & 56.25\% & -- \\
5.2 Disambiguation \& Total Count & 34.92\% & 0.5455 \\
5.3 Conditional \& Comparative Counting & 47.31\% & 0.8363 \\
\bottomrule
\end{tabular}
}
\end{table}

\subsection{Qwen2.5-Omni-3B}
\begin{table}[H]
\centering
\footnotesize
\caption{Performance Evaluation of Qwen2.5-Omni-3B}
\label{tab:qwen2.5-omni-3b}
\resizebox{\columnwidth}{!}{
\begin{tabular}{p{5cm}cc}
\toprule
\textbf{Category/Sub-category} & \textbf{MCQ Accuracy} & \textbf{QA Similarity} \\
\midrule
\textbf{Overall Performance} & 43.85\% & 0.6150 \\
\midrule
\textbf{1. Basic Understanding} & 38.89\% & 0.6296 \\
1.1 Perspective Definition \& Identification & 41.67\% & 0.4448 \\
1.2 Effect of Perspective on Object Shape & 46.59\% & 0.7782 \\
1.3 Perspective \& Occlusion & 29.69\% & 0.2898 \\
\midrule
\textbf{2. Image Characteristics} & 30.11\% & 0.6063 \\
2.1 Field of View \& Information & 32.29\% & 0.6257 \\
2.2 Distortion \& Perspective Effects & 27.50\% & 0.5556 \\
2.3 Multi-view Fusion & -- & 0.5973 \\
\midrule
\textbf{3. Perspective Question Design} & 43.56\% & 0.5537 \\
3.1 Position Inference & 44.74\% & 0.6198 \\
3.2 Spatial Relationships & 61.36\% & 0.6202 \\
3.3 Dynamic Perspective Change & 33.33\% & 0.4427 \\
3.4 Size Judgment & 43.43\% & -- \\
3.5 Perspective Transformation & 34.78\% & 0.6031 \\
\midrule
\textbf{4. Advanced Reasoning} & 72.31\% & 0.6627 \\
4.1 Multi-level Spatial Understanding & 54.55\% & 0.6894 \\
4.2 Inferring Implied Relationships & 77.03\% & 0.6581 \\
4.3 Inferring Scene Narrative & 76.62\% & 0.6081 \\
\midrule
\textbf{5. Quantitative Reasoning} & 35.17\% & 0.7950 \\
5.1 Occlusion-Based Counting & 37.50\% & -- \\
5.2 Disambiguation \& Total Count & 33.33\% & 0.7538 \\
5.3 Conditional \& Comparative Counting & 35.48\% & 0.8210 \\
\bottomrule
\end{tabular}
}
\end{table}

\subsection{Qwen2.5-VL-7B}
\begin{table}[H]
\centering
\footnotesize
\caption{Performance Evaluation of Qwen2.5-VL-7B}
\label{tab:qwen2.5-vl-7b}
\resizebox{\columnwidth}{!}{
\begin{tabular}{p{5cm}cc}
\toprule
\textbf{Category/Sub-category} & \textbf{MCQ Accuracy} & \textbf{QA Similarity} \\
\midrule
\textbf{Overall Performance} & 50.44\% & 0.7910 \\
\midrule
\textbf{1. Basic Understanding} & 39.35\% & 0.7838 \\
1.1 Perspective Definition \& Identification & 29.63\% & 0.7109 \\
1.2 Effect of Perspective on Object Shape & 52.27\% & 0.7932 \\
1.3 Perspective \& Occlusion & 46.88\% & 0.7654 \\
\midrule
\textbf{2. Image Characteristics} & 41.48\% & 0.7923 \\
2.1 Field of View \& Information & 41.67\% & 0.6675 \\
2.2 Distortion \& Perspective Effects & 41.25\% & 0.8668 \\
2.3 Multi-view Fusion & -- & 0.8672 \\
\midrule
\textbf{3. Perspective Question Design} & 62.87\% & 0.7743 \\
3.1 Position Inference & 68.42\% & 0.7487 \\
3.2 Spatial Relationships & 64.44\% & 0.8602 \\
3.3 Dynamic Perspective Change & 61.90\% & 0.7262 \\
3.4 Size Judgment & 66.67\% & -- \\
3.5 Perspective Transformation & 47.83\% & 0.7976 \\
\midrule
\textbf{4. Advanced Reasoning} & 73.33\% & 0.7583 \\
4.1 Multi-level Spatial Understanding & 72.73\% & 0.7496 \\
4.2 Inferring Implied Relationships & 85.14\% & 0.8892 \\
4.3 Inferring Scene Narrative & 62.34\% & 0.7693 \\
\midrule
\textbf{5. Quantitative Reasoning} & 37.71\% & 0.8798 \\
5.1 Occlusion-Based Counting & 38.75\% & -- \\
5.2 Disambiguation \& Total Count & 36.51\% & 0.8174 \\
5.3 Conditional \& Comparative Counting & 37.63\% & 0.9192 \\
\bottomrule
\end{tabular}
}
\end{table}

\subsection{Qwen2.5-VL-3B}
\begin{table}[H]
\centering
\footnotesize
\caption{Performance Evaluation of Qwen2.5-VL-3B}
\label{tab:qwen2.5-vl-3b}
\resizebox{\columnwidth}{!}{
\begin{tabular}{p{5cm}cc}
\toprule
\textbf{Category/Sub-category} & \textbf{MCQ Accuracy} & \textbf{QA Similarity} \\
\midrule
\textbf{Overall Performance} & 46.34\% & 0.7967 \\
\midrule
\textbf{1. Basic Understanding} & 40.74\% & 0.7975 \\
1.1 Perspective Definition \& Identification & 35.19\% & 0.7773 \\
1.2 Effect of Perspective on Object Shape & 59.09\% & 0.8125 \\
1.3 Perspective \& Occlusion & 37.50\% & 0.7632 \\
\midrule
\textbf{2. Image Characteristics} & 55.68\% & 0.7945 \\
2.1 Field of View \& Information & 47.92\% & 0.7210 \\
2.2 Distortion \& Perspective Effects & 65.00\% & 0.7823 \\
2.3 Multi-view Fusion & -- & 0.8424 \\
\midrule
\textbf{3. Perspective Question Design} & 47.52\% & 0.7895 \\
3.1 Position Inference & 39.47\% & 0.7506 \\
3.2 Spatial Relationships & 64.44\% & 0.8723 \\
3.3 Dynamic Perspective Change & 41.27\% & 0.7749 \\
3.4 Size Judgment & 57.58\% & -- \\
3.5 Perspective Transformation & 30.43\% & 0.7813 \\
\midrule
\textbf{4. Advanced Reasoning} & 55.38\% & 0.7634 \\
4.1 Multi-level Spatial Understanding & 34.09\% & 0.7554 \\
4.2 Inferring Implied Relationships & 72.97\% & 0.8141 \\
4.3 Inferring Scene Narrative & 50.65\% & 0.7772 \\
\midrule
\textbf{5. Quantitative Reasoning} & 36.02\% & 0.8512 \\
5.1 Occlusion-Based Counting & 31.25\% & -- \\
5.2 Disambiguation \& Total Count & 30.16\% & 0.7245 \\
5.3 Conditional \& Comparative Counting & 44.09\% & 0.9310 \\
\bottomrule
\end{tabular}
}
\end{table}

\subsection{Pano-R1-7B}
\begin{table}[H]
\centering
\footnotesize
\caption{Performance Evaluation of Pano-R1-7B}
\label{tab:qwen2.5-vl-7b-grpo}
\resizebox{\columnwidth}{!}{
\begin{tabular}{p{5cm}cc}
\toprule
\textbf{Category/Sub-category} & \textbf{MCQ Accuracy} & \textbf{QA Similarity} \\
\midrule
\textbf{Overall Performance} & 56.78\% & 0.8316 \\
\midrule
\textbf{1. Basic Understanding} & 42.13\% & 0.8255 \\
1.1 Perspective Definition \& Identification & 30.56\% & 0.9660 \\
1.2 Effect of Perspective on Object Shape & 65.91\% & 0.8380 \\
1.3 Perspective \& Occlusion & 45.31\% & 0.7892 \\
\midrule
\textbf{2. Image Characteristics} & 61.93\% & 0.8264 \\
2.1 Field of View \& Information & 67.71\% & 0.7370 \\
2.2 Distortion \& Perspective Effects & 55.00\% & 0.8693 \\
2.3 Multi-view Fusion & -- & 0.8807 \\
\midrule
\textbf{3. Perspective Question Design} & 65.35\% & 0.8304 \\
3.1 Position Inference & 78.95\% & 0.8003 \\
3.2 Spatial Relationships & 68.89\% & 0.8843 \\
3.3 Dynamic Perspective Change & 66.67\% & 0.8355 \\
3.4 Size Judgment & 63.64\% & -- \\
3.5 Perspective Transformation & 34.78\% & 0.8099 \\
\midrule
\textbf{4. Advanced Reasoning} & 76.41\% & 0.7545 \\
4.1 Multi-level Spatial Understanding & 68.18\% & 0.7400 \\
4.2 Inferring Implied Relationships & 90.54\% & 0.8730 \\
4.3 Inferring Scene Narrative & 67.53\% & 0.7780 \\
\midrule
\textbf{5. Quantitative Reasoning} & 42.80\% & 0.9110 \\
5.1 Occlusion-Based Counting & 45.00\% & -- \\
5.2 Disambiguation \& Total Count & 41.27\% & 0.8807 \\
5.3 Conditional \& Comparative Counting & 41.94\% & 0.9301 \\
\bottomrule
\end{tabular}
}
\end{table}

\subsection{Pano-R1-3B}
\begin{table}[H]
\centering
\footnotesize
\caption{Performance Evaluation of Pano-R1-3B}
\label{tab:qwen2.5-vl-3b-grpo}
\resizebox{\columnwidth}{!}{
\begin{tabular}{p{5cm}cc}
\toprule
\textbf{Category/Sub-category} & \textbf{MCQ Accuracy} & \textbf{QA Similarity} \\
\midrule
\textbf{Overall Performance} & 49.66\% & 0.8037 \\
\midrule
\textbf{1. Basic Understanding} & 44.91\% & 0.7893 \\
1.1 Perspective Definition \& Identification & 42.59\% & 0.9159 \\
1.2 Effect of Perspective on Object Shape & 65.91\% & 0.7964 \\
1.3 Perspective \& Occlusion & 34.38\% & 0.7680 \\
\midrule
\textbf{2. Image Characteristics} & 55.11\% & 0.7979 \\
2.1 Field of View \& Information & 55.21\% & 0.7682 \\
2.2 Distortion \& Perspective Effects & 55.00\% & 0.8481 \\
2.3 Multi-view Fusion & -- & 0.8130 \\
\midrule
\textbf{3. Perspective Question Design} & 48.51\% & 0.8054 \\
3.1 Position Inference & 50.00\% & 0.7645 \\
3.2 Spatial Relationships & 66.67\% & 0.8463 \\
3.3 Dynamic Perspective Change & 39.68\% & 0.8320 \\
3.4 Size Judgment & 60.61\% & -- \\
3.5 Perspective Transformation & 17.39\% & 0.7703 \\
\midrule
\textbf{4. Advanced Reasoning} & 64.10\% & 0.7531 \\
4.1 Multi-level Spatial Understanding & 50.00\% & 0.7493 \\
4.2 Inferring Implied Relationships & 83.78\% & 0.7605 \\
4.3 Inferring Scene Narrative & 53.25\% & 0.7600 \\
\midrule
\textbf{5. Quantitative Reasoning} & 38.98\% & 0.8639 \\
5.1 Occlusion-Based Counting & 42.50\% & -- \\
5.2 Disambiguation \& Total Count & 28.57\% & 0.8256 \\
5.3 Conditional \& Comparative Counting & 43.01\% & 0.8881 \\
\bottomrule
\end{tabular}
}
\end{table}

\section{Dataset Information}
This appendix provides more detailed information about the datasets we introduced.

\begin{table}[H]
\centering
\caption{CFPano VQA Dataset Overview}
\label{tab:dataset_overview}
\footnotesize
\begin{tabular}{lccc}
\toprule
\textbf{Dataset} & \textbf{Total Questions} & \textbf{MCQ Count (\%)} & \textbf{QA Count (\%)} \\
\midrule
Overall Dataset & 8,094 & 5,240 (64.74\%) & 2,854 (35.26\%) \\
Training Set & 6,475 & 4,215 (65.10\%) & 2,260 (34.90\%) \\
Testing Set & 1,619 & 1,025 (63.31\%) & 594 (36.69\%) \\
\bottomrule
\end{tabular}
\end{table}

\begin{table}[H]
\centering
\caption{Overall Dataset Category Distribution}
\label{tab:overall_distribution}
\scriptsize
\resizebox{\columnwidth}{!}{
\begin{tabular}{p{6cm}cc}
\toprule
\textbf{Category/Sub-category} & \textbf{Count} & \textbf{Percentage} \\
\midrule
\textbf{Total Questions} & \textbf{8,094} & \textbf{100.00\%} \\
\midrule
\textbf{1. Basic Understanding} & 1,534 & 18.95\% \\
1.1 Perspective Definition \& Identification & 569 & 7.03\% \\
1.2 Effect of Perspective on Object Shape & 500 & 6.18\% \\
1.3 Perspective \& Occlusion & 465 & 5.74\% \\
\midrule
\textbf{2. Image Characteristics} & 1,466 & 18.11\% \\
2.1 Field of View \& Information & 670 & 8.28\% \\
2.2 Distortion \& Perspective Effects & 465 & 5.74\% \\
2.3 Multi-view Fusion & 331 & 4.09\% \\
\midrule
\textbf{3. Perspective Question Design} & 2,331 & 28.80\% \\
3.1 Position Inference & 450 & 5.56\% \\
3.2 Spatial Relationships & 452 & 5.58\% \\
3.3 Dynamic Perspective Change & 723 & 8.93\% \\
3.4 Size Judgment & 231 & 2.85\% \\
3.5 Perspective Transformation & 475 & 5.87\% \\
\midrule
\textbf{4. Advanced Reasoning} & 1,223 & 15.11\% \\
4.1 Multi-level Spatial Understanding & 450 & 5.56\% \\
4.2 Inferring Implied Relationships & 322 & 3.98\% \\
4.3 Inferring Scene Narrative & 451 & 5.57\% \\
\midrule
\textbf{5. Quantitative Reasoning} & 1,540 & 19.03\% \\
5.1 Occlusion-Based Counting & 384 & 4.74\% \\
5.2 Disambiguation \& Total Count & 481 & 5.94\% \\
5.3 Conditional \& Comparative Counting & 675 & 8.34\% \\
\bottomrule
\end{tabular}
}
\end{table}

\begin{table}[H]
\centering
\caption{Training Set Category Distribution}
\label{tab:train_distribution}
\scriptsize
\resizebox{\columnwidth}{!}{
\begin{tabular}{p{6cm}cc}
\toprule
\textbf{Category/Sub-category} & \textbf{Count} & \textbf{Percentage} \\
\midrule
\textbf{Total Training Questions} & \textbf{6,475} & \textbf{100.00\%} \\
\midrule
\textbf{1. Basic Understanding} & 1,250 & 19.31\% \\
1.1 Perspective Definition \& Identification & 460 & 7.10\% \\
1.2 Effect of Perspective on Object Shape & 409 & 6.32\% \\
1.3 Perspective \& Occlusion & 381 & 5.88\% \\
\midrule
\textbf{2. Image Characteristics} & 1,162 & 17.95\% \\
2.1 Field of View \& Information & 526 & 8.12\% \\
2.2 Distortion \& Perspective Effects & 380 & 5.87\% \\
2.3 Multi-view Fusion & 256 & 3.95\% \\
\midrule
\textbf{3. Perspective Question Design} & 1,865 & 28.80\% \\
3.1 Position Inference & 359 & 5.54\% \\
3.2 Spatial Relationships & 359 & 5.54\% \\
3.3 Dynamic Perspective Change & 568 & 8.77\% \\
3.4 Size Judgment & 198 & 3.06\% \\
3.5 Perspective Transformation & 381 & 5.88\% \\
\midrule
\textbf{4. Advanced Reasoning} & 969 & 14.97\% \\
4.1 Multi-level Spatial Understanding & 367 & 5.67\% \\
4.2 Inferring Implied Relationships & 247 & 3.81\% \\
4.3 Inferring Scene Narrative & 355 & 5.48\% \\
\midrule
\textbf{5. Quantitative Reasoning} & 1,229 & 18.98\% \\
5.1 Occlusion-Based Counting & 304 & 4.69\% \\
5.2 Disambiguation \& Total Count & 389 & 6.01\% \\
5.3 Conditional \& Comparative Counting & 536 & 8.28\% \\
\bottomrule
\end{tabular}
}
\end{table}

\begin{table}[ht]
\centering
\caption{Testing Set Category Distribution}
\label{tab:test_distribution}
\scriptsize
\resizebox{\columnwidth}{!}{
\begin{tabular}{p{6cm}cc}
\toprule
\textbf{Category/Sub-category} & \textbf{Count} & \textbf{Percentage} \\
\midrule
\textbf{Total Testing Questions} & \textbf{1,619} & \textbf{100.00\%} \\
\midrule
\textbf{1. Basic Understanding} & 284 & 17.54\% \\
1.1 Perspective Definition \& Identification & 109 & 6.73\% \\
1.2 Effect of Perspective on Object Shape & 91 & 5.62\% \\
1.3 Perspective \& Occlusion & 84 & 5.19\% \\
\midrule
\textbf{2. Image Characteristics} & 304 & 18.78\% \\
2.1 Field of View \& Information & 144 & 8.89\% \\
2.2 Distortion \& Perspective Effects & 85 & 5.25\% \\
2.3 Multi-view Fusion & 75 & 4.63\% \\
\midrule
\textbf{3. Perspective Question Design} & 466 & 28.78\% \\
3.1 Position Inference & 91 & 5.62\% \\
3.2 Spatial Relationships & 93 & 5.74\% \\
3.3 Dynamic Perspective Change & 155 & 9.57\% \\
3.4 Size Judgment & 33 & 2.04\% \\
3.5 Perspective Transformation & 94 & 5.81\% \\
\midrule
\textbf{4. Advanced Reasoning} & 254 & 15.69\% \\
4.1 Multi-level Spatial Understanding & 83 & 5.13\% \\
4.2 Inferring Implied Relationships & 75 & 4.63\% \\
4.3 Inferring Scene Narrative & 96 & 5.93\% \\
\midrule
\textbf{5. Quantitative Reasoning} & 311 & 19.21\% \\
5.1 Occlusion-Based Counting & 80 & 4.94\% \\
5.2 Disambiguation \& Total Count & 92 & 5.68\% \\
5.3 Conditional \& Comparative Counting & 139 & 8.59\% \\
\bottomrule
\end{tabular}
}
\end{table}

\section{Question Templates}
This appendix includes a large table of question templates. Flex means both question type are applied in this question tempate text.

\begin{table*}[ht]
\centering
\caption{Complete Question Templates for ReplicaPano VQA Dataset}
\label{tab:question_templates}
\tiny
\begin{tabular}{p{0.8cm}p{3.5cm}p{0.6cm}p{0.4cm}p{11cm}}
\toprule
\textbf{Main Category} & \textbf{Sub-Category} & \textbf{Template ID} & \textbf{Type} & \textbf{Question Template Text} \\
\midrule
\multirow{6}{0.8cm}{1. Basic Understanding} & \multirow{2}{3.5cm}{1.1 Perspective Definition \& Identification} & T1.1.1 & MCQ & Comparing \texttt{\{frame\_X\}} and \texttt{\{frame\_Y\}}, which one was taken from a higher viewpoint? \\
& & T1.1.2 & MCQ & Which frame, \texttt{\{frame\_X\}} or \texttt{\{frame\_Y\}}, provides a more direct top-down view of \texttt{\{object\_A\}}? \\
\cmidrule{2-5}
& \multirow{2}{3.5cm}{1.2 Effect of Perspective on Object Shape} & T1.2.1 & QA & Describe how the apparent shape of \texttt{\{object\_A\}} changes between the perspective of \texttt{\{frame\_X\}} and \texttt{\{frame\_Y\}}. \\
& & T1.2.2 & MCQ & Which frame, \texttt{\{frame\_X\}} or \texttt{\{frame\_Y\}}, provides a more comprehensive view of the three-dimensional shape of \texttt{\{object\_A\}}? \\
\cmidrule{2-5}
& \multirow{2}{3.5cm}{1.3 Perspective \& Occlusion} & T1.3.1 & Flex & In \texttt{\{frame\_X\}}, \texttt{\{object\_A\}} is partially occluded. Based on the view in \texttt{\{frame\_Y\}}, identify the object that is causing the occlusion. \\
& & T1.3.2 & MCQ & By comparing the two frames, which perspective reveals more about the spatial relationship (front/back) between \texttt{\{object\_A\}} and \texttt{\{object\_B\}}? \\
\midrule
\multirow{7}{0.8cm}{2. Image Characteristics} & \multirow{3}{3.5cm}{2.1 Field of View \& Information} & T2.1.1 & MCQ & Comparing the two images, which one (\texttt{\{frame\_X\}} or \texttt{\{frame\_Y\}}) captures a wider field of view of the scene? \\
& & T2.1.2 & MCQ & Considering an object located at the edge of the view in \texttt{\{frame\_X\}}, does this same object appear closer to the center in \texttt{\{frame\_Y\}}? \\
& & T2.1.3 & QA & What does it imply about the camera's movement if an object appears to move from the edge of the view toward the center? \\
\cmidrule{2-5}
& \multirow{2}{3.5cm}{2.2 Distortion \& Perspective Effects} & T2.2.1 & MCQ & In which frame is the perspective effect, where distant objects appear smaller, more pronounced? \\
& & T2.2.2 & MCQ & In \texttt{\{frame\_X\}}, \texttt{\{object\_A\}} and \texttt{\{object\_B\}} appear to be similar in size. Using the information from \texttt{\{frame\_Y\}}, are their actual sizes likely to be the same? \\
\cmidrule{2-5}
& \multirow{2}{3.5cm}{2.3 Multi-view Fusion} & T2.3.1 & Flex & The depth relationship between \texttt{\{object\_A\}} and \texttt{\{object\_B\}} is ambiguous in \texttt{\{frame\_X\}}. What visual cue does \texttt{\{frame\_Y\}} provide to resolve this ambiguity? \\
& & T2.3.2 & QA & By combining information from both views, describe the approximate location of \texttt{\{object\_A\}} within the room (e.g., near a corner, in the center). \\
\midrule
\multirow{10}{0.8cm}{3. Perspective Question Design} & \multirow{2}{3.5cm}{3.1 Position Inference} & T3.1.1 & QA & \texttt{\{frame\_X\}} shows the front of \texttt{\{object\_A\}}, while \texttt{\{frame\_Y\}} shows its side. Describe the camera's direction of movement from \texttt{\{frame\_X\}} to \texttt{\{frame\_Y\}}. \\
& & T3.1.2 & MCQ & Using both views, determine if \texttt{\{object\_A\}} is freestanding or placed against a wall or another object. \\
\cmidrule{2-5}
& \multirow{2}{3.5cm}{3.2 Spatial Relationships} & T3.2.1 & MCQ & By comparing the two perspectives, determine if \texttt{\{object\_A\}} is in front of or behind \texttt{\{object\_B\}}. \\
& & T3.2.2 & QA & In \texttt{\{frame\_X\}}, \texttt{\{object\_A\}} is to the left of \texttt{\{object\_B\}}, but in \texttt{\{frame\_Y\}}, it is on the right. Explain why this change in relative position occurred. \\
\cmidrule{2-5}
& \multirow{3}{3.5cm}{3.3 Dynamic Perspective Change} & T3.3.1 & QA & Describe the type of camera motion between \texttt{\{frame\_X\}} and \texttt{\{frame\_Y\}} (e.g., was it a pan, a rotation, or a zoom?). \\
& & T3.3.2 & MCQ & As the perspective changes from \texttt{\{frame\_X\}} to \texttt{\{frame\_Y\}}, which object's apparent position shifts the most within the frame? \\
& & T3.3.3 & QA & What does a larger apparent shift in an object's position (parallax) indicate about its relative distance from the camera? \\
\cmidrule{2-5}
& \multirow{2}{3.5cm}{3.4 Size Judgment} & T3.4.1 & QA & Explain why \texttt{\{object\_A\}} appears significantly larger or smaller in \texttt{\{frame\_X\}} compared to \texttt{\{frame\_Y\}}. \\
& & T3.4.2 & MCQ & Based on the visual evidence in both frames, which object, \texttt{\{object\_A\}} or \texttt{\{object\_B\}}, is likely larger in reality? \\
\cmidrule{2-5}
& \multirow{2}{3.5cm}{3.5 Perspective Transformation} & T3.5.1 & Flex & If you were to take a third picture to get a clear view of the occluded part of \texttt{\{object\_A\}}, where should the camera be moved to? \\
& & T3.5.2 & QA & Based on the visible parts of \texttt{\{object\_A\}} in both frames, describe its likely appearance if viewed from directly above. \\
\midrule
\multirow{6}{0.8cm}{4. Advanced Reasoning} & \multirow{2}{3.5cm}{4.1 Multi-level Spatial Understanding} & T4.1.1 & QA & Using both frames, describe the foreground, mid-ground, and background elements of this scene. \\
& & T4.1.2 & MCQ & Using evidence from both frames, determine if \texttt{\{object\_A\}} is closer to \texttt{\{object\_B\}} or to \texttt{\{object\_C\}}. \\
\cmidrule{2-5}
& \multirow{2}{3.5cm}{4.2 Inferring Implied Relationships} & T4.2.1 & MCQ & Given that \texttt{\{object\_A\}} and \texttt{\{object\_B\}} are positioned close to each other in both views, what is their likely functional relationship (e.g., a chair and a table)? \\
& & T4.2.2 & MCQ & Based on the arrangement of objects visible across both frames, what do you infer is the primary use of this room (e.g., office, bedroom)? \\
\cmidrule{2-5}
& \multirow{2}{3.5cm}{4.3 Inferring Scene Narrative} & T4.3.1 & MCQ & Based on the layout derived from both frames, is there a clear, unobstructed path for someone to walk from \texttt{\{object\_A\}} to \texttt{\{object\_B\}}? \\
& & T4.3.2 & Flex & If a light source were placed directly above \texttt{\{object\_A\}}, which surfaces of \texttt{\{object\_B\}} would be illuminated, based on their spatial relationship? \\
\midrule
\multirow{7}{0.8cm}{5. Quantitative Reasoning} & \multirow{2}{3.5cm}{5.1 Occlusion-Based Counting} & T5.1.1 & MCQ & In \texttt{\{frame\_X\}}, some \texttt{\{object\_type\_plural\}} are hidden. Using \texttt{\{frame\_Y\}} to see the full scene, what is the true total number of \texttt{\{object\_type\_plural\}}? \\
& & T5.1.2 & MCQ & \texttt{\{frame\_Y\}} provides a clearer view of an area that is partially obscured in \texttt{\{frame\_X\}}. How many \texttt{\{object\_type\_plural\}} are revealed in that specific area by the change in perspective? \\
\cmidrule{2-5}
& \multirow{2}{3.5cm}{5.2 Disambiguation \& Total Count} & T5.2.1 & MCQ & By carefully examining both \texttt{\{frame\_X\}} and \texttt{\{frame\_Y\}}, determine the total count of distinct \texttt{\{object\_type\_plural\}} in the entire scene. \\
& & T5.2.2 & Flex & In \texttt{\{frame\_X\}}, it is unclear if \texttt{\{object\_A\}} is one or two separate items. Which frame clarifies this, and what is the correct count? \\
\cmidrule{2-5}
& \multirow{2}{3.5cm}{5.3 Conditional \& Comparative Counting} & T5.3.1 & MCQ & Based on the spatial layout visible across both frames, how many \texttt{\{object\_type\_plural\}} are located on top of \texttt{\{surface\_object\}}? \\
& & T5.3.2 & MCQ & Considering the entire scene visible in both frames, are there more \texttt{\{object\_type\_A\_plural\}} or \texttt{\{object\_type\_B\_plural\}}? \\
\bottomrule
\end{tabular}
\end{table*}

\section{Model Inference Cases}
This appendix presents three model inference cases, each in a large figure spanning two columns. Each case compares the performance of Qwen2.5VL-7B-Instruct, InternVL3-9B, and Pano-R1 on panoramic images, highlighting differences in reasoning and output accuracy.

% \begin{figure*}[ht] % 双栏大图
% \centering
% \includegraphics[width=\textwidth]{Figures/Case1_cropped.pdf} % 替换为您的图像文件
% \caption{Model Inference Case 1: Qwen2.5VL-7B-Instruct's reasoning content is partially correct, but the answer is completely wrong. InternVL3-9B's reasoning and answer are both wrong. Pano-R1's reasoning and answer are both correct.}
% \label{fig:inference-case1}
% \end{figure*}

\begin{figure*}[ht] % Double-column wide figure
\centering
\includegraphics[width=\textwidth]{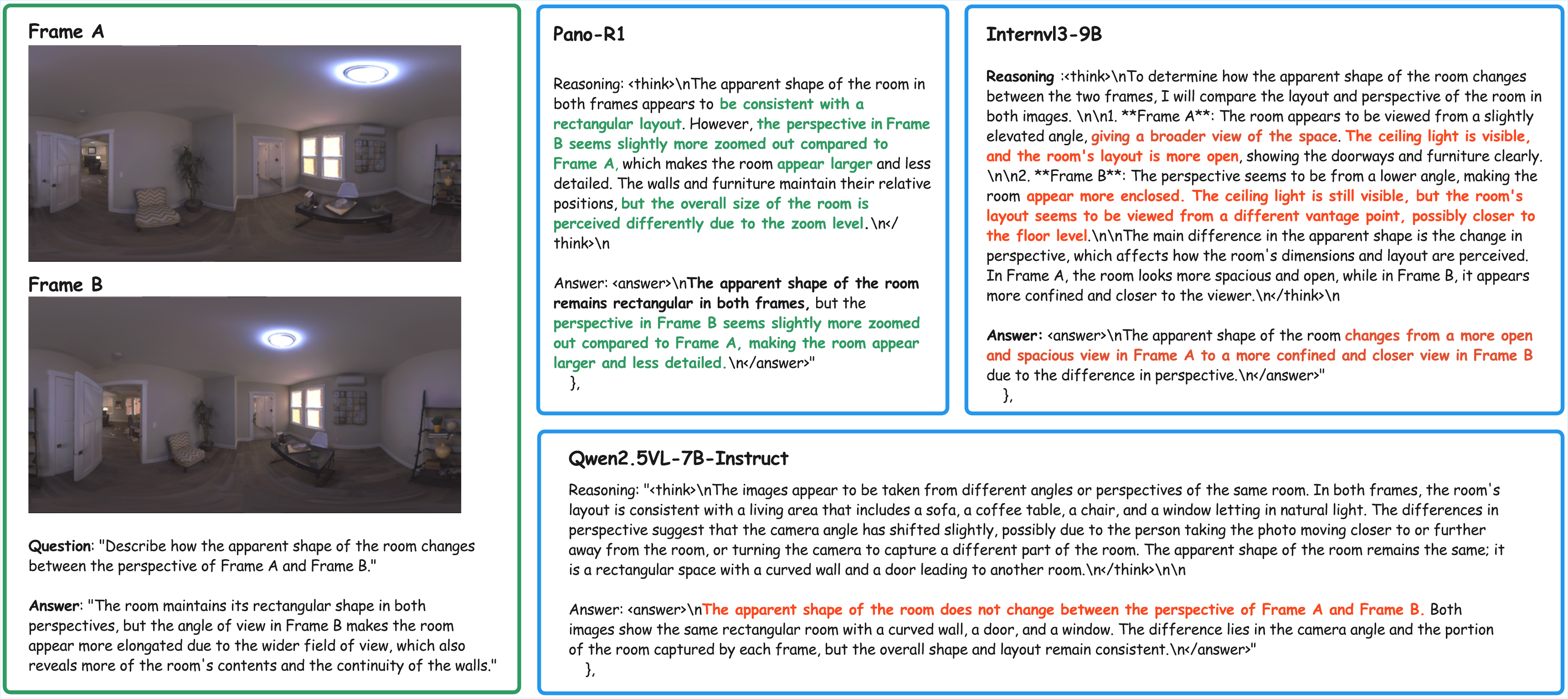} % Replace with your image file
\caption{The second case of model inference between our model (Pano-R1) and other leading MLLMs.}
\Description{This figure presents a qualitative analysis of three models on a nuanced visual reasoning task. The models are asked to describe the change in the apparent shape of a room between two slightly different perspectives, Frame A and Frame B. Our model, Pano-R1, correctly identifies that the room's overall shape remains rectangular but astutely notes the subtle difference in perspective, where Frame B seems slightly more zoomed out. Its reasoning and answer are both accurate and consistent. In contrast, Qwen2.5VL-7B-Instruct provides partially correct reasoning about camera shifts but delivers a completely wrong answer. InternVL3-9B demonstrates a fundamental misunderstanding of the scene, with both its reasoning and final answer being incorrect. This case highlights Pano-R1's superior ability to discern subtle visual details and provide accurate, coherent descriptions.}
\label{fig:inference-case1}
\end{figure*}

% \begin{figure*}[ht]
% \centering
% \includegraphics[width=\textwidth]{Figures/Case2_cropped.pdf}
% \caption{The third case of model inference between our model (Pano-R1) and other leading MLLMs}
% % \Description{An image showing counted objects and step-by-step reasoning annotations. For Qwen2.5VL-7B-Instruct, both reasoning and answer are wrong, and it failed to output in the specified format. For InternVL3-9B and Pano-R1, both reasoning and answer are correct.}
% \label{fig:inference-case2}
% \end{figure*}

\begin{figure*}[ht]
\centering
\includegraphics[width=\textwidth]{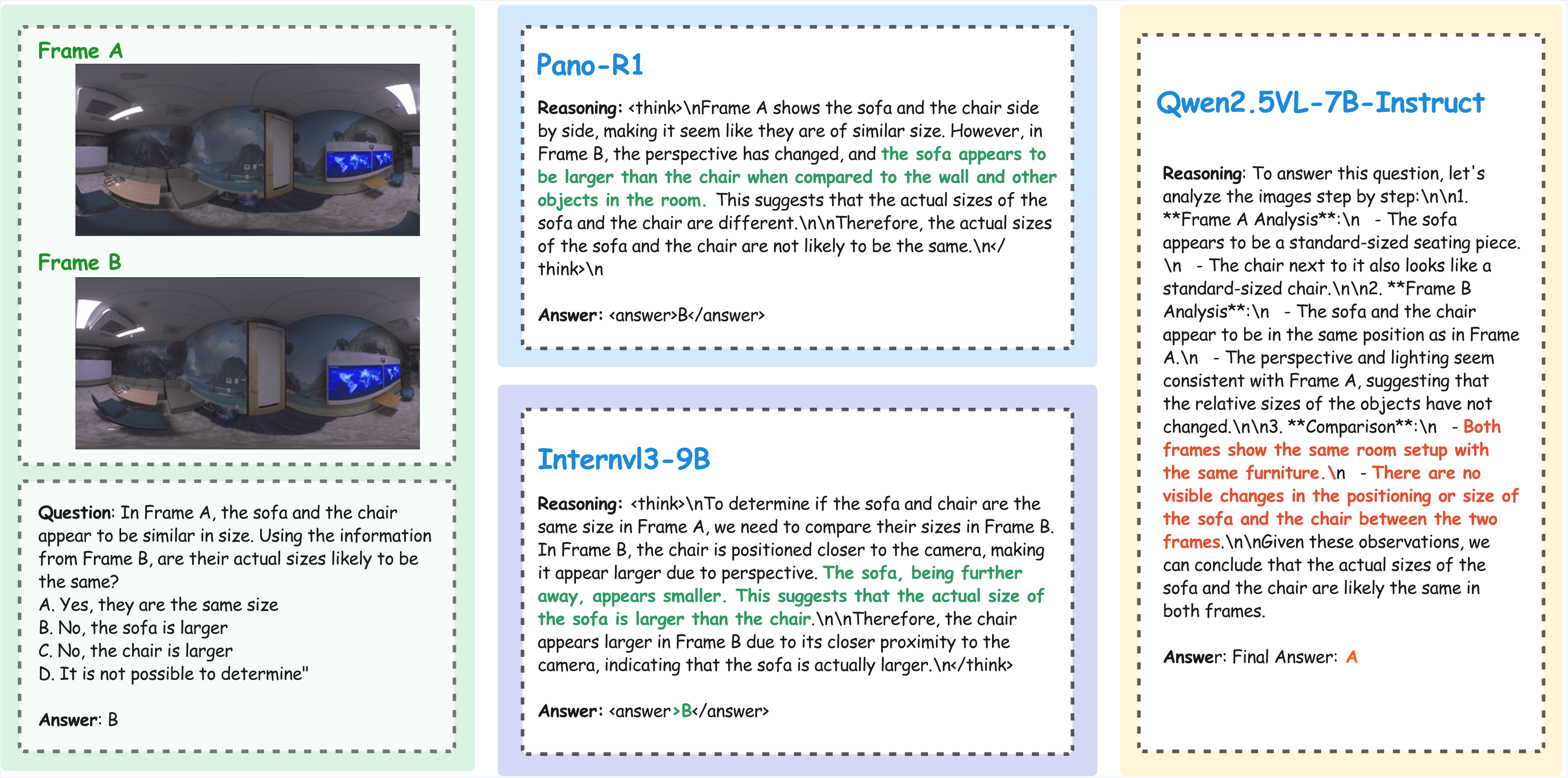}
\caption{The third case of model inference between our model (Pano-R1) and other leading MLLMs}
\Description{This figure illustrates a qualitative comparison on a challenging inference task: determining the actual size relationship between a sofa and a chair using two different visual perspectives. In Frame A, the objects appear similarly sized. Frame B provides a different viewpoint. Our model, Pano-R1, and InternVL3-9B both correctly interpret the perspective shift in Frame B, reasoning that changes in apparent size indicate the sofa is actually larger. In contrast, Qwen2.5VL-7B-Instruct fails to perceive any change, incorrectly concluding the objects are the same size; its reasoning and answer are both wrong, and it also fails to adhere to the specified output format. This case demonstrates the superior ability of Pano-R1 and InternVL3-9B to perform complex spatial reasoning by integrating information across multiple views.}
\label{fig:inference-case2}
\end{figure*}

\end{document}